\documentclass[letterpaper, 10 pt, conference]{ieeeconf}  

\IEEEoverridecommandlockouts                              

\usepackage{graphicx}
\usepackage{caption}
\usepackage{times} 
\usepackage{amsmath} 
\usepackage{amssymb}  
\usepackage{mathtools}
\usepackage{amsfonts}
\usepackage[usenames,dvipsnames,svgnames,table, xcdraw]{xcolor}
\usepackage[colorlinks=true,pdfpagemode=UseNone,citecolor=black,linkcolor=black,urlcolor=BrickRed]{hyperref}
\usepackage[ruled,vlined,linesnumbered]{algorithm2e}
\SetKw{Continue}{continue}
\SetKw{Break}{break}

\usepackage{cite} 
\usepackage{url}
\usepackage{subcaption}
\usepackage{booktabs}
\newcommand{\squeezeup}{\vspace{-3mm}}

\usepackage{multirow}

\newcommand{\transpose}{^\mathsf{T}}

\title{ \LARGE \bf
Heterogeneous Vehicle Routing and Teaming with Gaussian Distributed Energy Uncertainty
}

\author{Bo Fu$^{1}$, William Smith$^{2}$, Denise Rizzo$^{2}$, Matthew Castanier$^{2}$, and Kira Barton$^{1}$
\thanks{DISTRIBUTION A. Approved for public release; distribution unlimited. (OPSEC 4396)} 
\thanks{$^{1}$ Bo Fu and Kira Barton are with the University of Michigan, Ann Arbor, MI 48109, USA {\tt \small \{bofu, bartonkl\}@umich.edu}}
\thanks{$^{2}$ William Smith, Denise Rizzo, and Matthew Castanier are with the US Army CCDC Ground Vehicle Systems Center, Warren, MI 48397, USA \newline {\tt \small \{william.c.smith1019.civ, denise.m.rizzo2.civ, matthew.p.castanier.civ\}@mail.mil}}%
}

\begin{document}

\maketitle

\begin{abstract}
For robot swarms operating on complex missions in an uncertain environment, it is important that the decision-making algorithm considers both heterogeneity and uncertainty. This paper presents a stochastic programming framework for the vehicle routing problem with stochastic travel energy costs and heterogeneous vehicles and tasks. We represent the heterogeneity as linear constraints, estimate the uncertain energy cost through Gaussian process regression, formulate this stochasticity as chance constraints or stochastic recourse costs, and then solve the stochastic programs using branch and cut algorithms to minimize the expected energy cost. The performance and practicality are demonstrated through extensive computational experiments and a practical test case.
\end{abstract}



\section{Introduction and Related Work}\label{sec:introduction}
Recent advances in robotic sensing and control have enabled the application of robotic systems in a variety of unstructured natural environments, including robotic agriculture, underwater exploration, and search and rescue in caves. In the meantime, the progress in multi-robot decision-making algorithms and the growth in computational power allow robot swarms to work together to complete missions. Despite the achievements made by these decision-making algorithms, most of them are applied to a team of homogeneous robots for one specific task and require a structured known environment. As robotic systems start to operate on more complex missions, e.g. a military mission that contains several scout, breach, and delivery tasks at different locations, it is important to consider uncertainty in the environment and heterogeneity of the robots and tasks.

\begin{figure}[t!]
	\centering
	\includegraphics[width=0.9\linewidth]{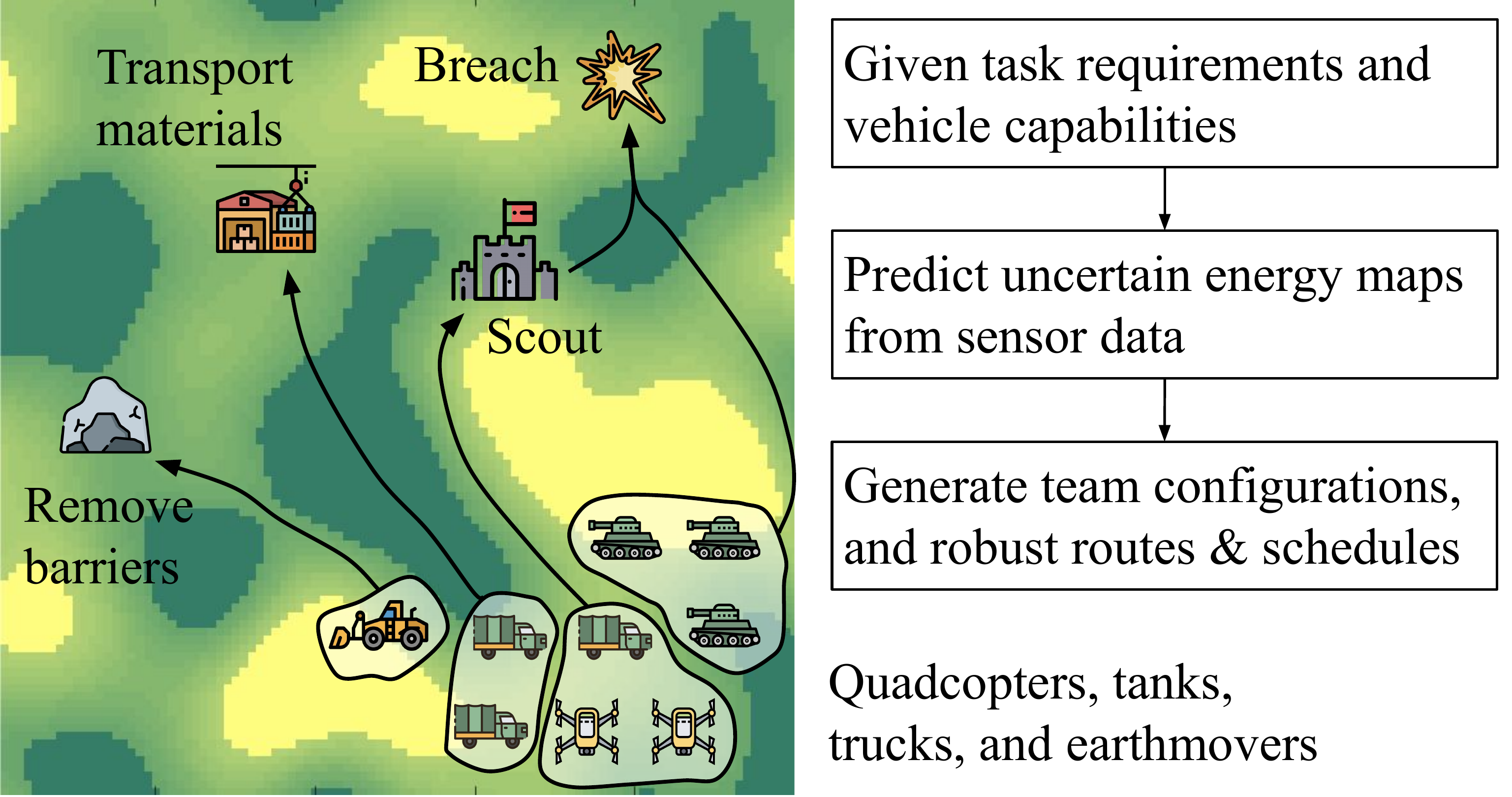}
	\caption{An example for a SHTP with travel energy uncertainty, where there are 4 tasks to be completed and 9 vehicles for selection.}
	\label{fig:teaming_problem_overview}
    \vspace{-0.1cm}
\end{figure}

We define this class of problems as the stochastic heterogeneous teaming problem (SHTP): given a set of heterogeneous vehicles (robots) and heterogeneous tasks distributed within an uncertain environment (a part of the information is partially known), form vehicle teams to complete the tasks and optimize an objective function considering the uncertainty. In this work, the uncertainty considered is in the travel energy cost, and an example problem is shown in Fig. \ref{fig:teaming_problem_overview}.

If the environment is fully known and there is only one type of vehicle and task, SHTP reduces to a vehicle routing problem (VRP): find the optimal routes for a fleet of homogeneous vehicles to visit a set of task targets, satisfy the demand of all targets, and minimize some objective function. To address the uncertain nature of real-world applications, a part of the parameters in the VRP are modeled as random variables, and the problem turns into a stochastic vehicle routing problem (SVRP). 
The types of uncertainty studied in previous work include uncertain demand value at each target \cite{laporte2002integer,mendoza2013multi,secomandi2009reoptimization,marinakis2013particle}
, uncertain service time for a target \cite{sundar2017path,chen2014optimizing,li2010vehicle,gomez2016modeling}, uncertain travel time between two targets \cite{gomez2016modeling,chen2014optimizing,li2010vehicle,russell2008vehicle},
, and uncertain travel energy costs \cite{venkatachalam2019two,venkatachalam2018two}. These SVRPs are typically formulated as integer programs, and solved through either exact algorithms (e.g. branch and cut  \cite{sundar2017path,laporte2002integer,chen2014optimizing,venkatachalam2019two}), which returns the optimal solution, or heuristic algorithms (e.g. adaptive large neighborhood search \cite{chen2014optimizing}, tabu search \cite{li2010vehicle,venkatachalam2018two,haugland2007designing,solano2015local}, or constructive algorithms \cite{mendoza2013multi,venkatachalam2018two}).

Travel energy cost uncertainty exists widely in off-road operations; however, few previous works in SVRP consider energy uncertainty or provide theoretical support for their assumed energy distribution. The authors in \cite{venkatachalam2019two} propose a stochastic program 
where the fuel consumption for traveling between tasks is stochastic. The distribution of fuel consumption is represented by a number of random vector realizations with the associated probability. This non-parametric representation allows their model to stay linear, but makes the program large due to the number of realizations needed to represent the distribution. Instead, we select a nonlinear parametric modeling framework to limit the program size.

Heterogeneity can be classified into two categories \cite{sundar2017path}, structural and functional. Structural heterogeneity usually leads to less phenomenal differences that can be described as parameters, e.g. maximum velocity or energy capacity. Examples of functional heterogeneity include the ability to fly or breach enemies. The heterogeneity described in SHTP refers to functional heterogeneity, which can lead to fundamental differences between task capabilities. Few previous works address this heterogeneity.

In this paper, we leverage our previous work in \cite{quann2017energy,quann2018ground,quann2019chance} to model travel energy cost as Gaussian distributions. The uncertain energy cost is then combined with heterogeneous vehicles and tasks to formulate two stochastic mixed integer nonlinear programming models (MINLP). Two branch and cut algorithms are used to solve the MINLPs.

Contributions of this work include:
\begin{itemize}
    \item Development of a flexible framework to encode vehicle and task heterogeneity, which can be used for a broad set of practical problems.
    \item Derivation of SVRP models that incorporate Gaussian distributed energy costs.
    \item Evaluation of the proposed framework through a comprehensive computational investigation and candidate case study.
\end{itemize}


\section{Problem and Optimization Models}\label{sec:optimization_model}

In this section, we demonstrate how to encode heterogeneity and energy uncertainty within the constraints and objective function, to formulate the SHTP as an MINLP.

\subsection{Problem Description}\label{sec:problem_description}

Consider a set of vehicles \(V = \{v_1,\cdots,v_{n_v}\}\), capabilities \(A = \{a_1,\cdots,a_{n_a}\}\), and tasks \(M = \{m_1, \cdots, m_{n_m}\}\). Each vehicle \(k \in V\) is associated with a non-negative capability vector \(\mathbf{c}_k = [c_{k a_1}, \cdots, c_{k a_{n_a}}]\transpose\) (examples are in TABLE \ref{tab:vehicle_capability}). Each task \(i \in M\) requires a vehicle team with appropriate capabilities that drives the task requirement function \(\rho_i(\cdot)\) to 1. The goal is to determine the optimal task schedule for a set of vehicles, such that, all tasks are completed, the travel costs do not exceed the vehicle energy capacities, and the expected sum energy is minimized. It is assumed that a task can only occur when all vehicles in a team have arrived, introducing additional time scheduling constraints.

A task requirement function is a binary function of similar structure as (\ref{eqn:task_requirement_function}). \(\geq\), \(\land\), and \(\lor\) are ``greater than", ``and", and ``or" logical operators that return 1 if their conditions are satisfied, and 0 otherwise. \(\alpha_a\) is the sum value of capability \(a\) of the vehicles in the team as in (\ref{eqn:task_input}). \(\gamma_a\) is a predefined number reflecting the task requirement on capability \(a\). Note that (\ref{eqn:task_requirement_function}) is just an example, and there can be an arbitrary number of \(\land\) and \(\lor\) operations. {\color{black}The \(\geq\) can also be \(\leq\).} For more examples, refer to Sec. \ref{sec:practical_problem_application}.
\begin{align}
    \rho_{m_1}(\alpha_{a_1}, \alpha_{a_2}, &\cdots) = [(\alpha_{a_2} \geq \gamma_{a_2}) \lor (\alpha_{a_3} \geq \gamma_{a_3})] \land \nonumber\\
    [\alpha_{a_4} \geq &\gamma_{a_4}] \land[(\alpha_{a_5} \geq \gamma_{a_5}) \lor (\alpha_{a_6} \geq \gamma_{a_6})] \label{eqn:task_requirement_function} \\
    \alpha_a = \underset{k \in V_{m_1}}{\sum} c_{ka}&, \quad  \forall a \in A \ (V_{m_1} \text{ is the team for } m_1) \label{eqn:task_input}
\end{align}
For some non-additive capability types (such as maximum velocity), the team's capability is not the sum, but maximum, minimum, or other more complicated functions of individuals' capabilities, which require further modeling. In this work, we assume the capabilities are additive.

\subsection{Deterministic Model}
\begin{figure}[hbt!]
	\centering
	\includegraphics[width=0.8\linewidth]{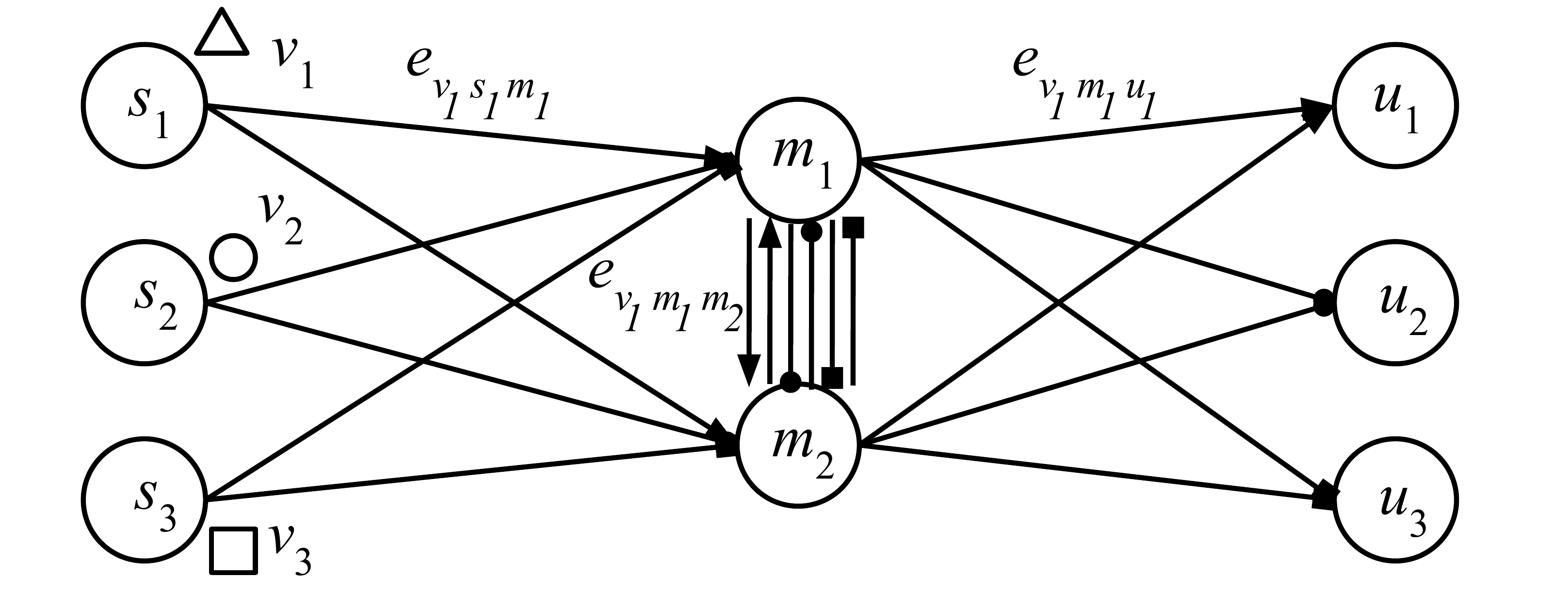}
	\caption{The graphical model with 3 vehicles and 2 tasks. The arrow type differentiates which vehicle an edge belongs to. Three edges are marked as examples.}
	\label{fig:graphical_model}
\vspace{-0.2cm}
\end{figure}

As shown in Fig. \ref{fig:graphical_model}, we first define a directed graph \(G = (S \cup U \cup M,E)\), with a set of vertices \(S \cup U \cup M\) and edges \(E\). \(S = \{s_1, \cdots, s_{n_v}\}\) and \(U = \{u_1, \cdots, u_{n_v}\}\) are the sets of start and terminal nodes respectively for each vehicle \(v_i \in V\). \(M = \{m_1, \cdots, m_{n_m}\}\) is the set of the task nodes.

For each vehicle \(v_k \in V\), there is an edge \(e_{v_k s_k i}\) from the start node to the task node \(i \in M\) and an edge \(e_{v_k i u_k}\) from task node \(i \in M\) to its terminal node. For simplicity, we also use notation \(e_{k s i}\) for \(e_{v_k s_k i}\) and \(e_{k i u}\) for \(e_{v_k i u_k}\), as there will be no ambiguity for the start and terminal node once \(v_k\) is determined. Between each task node pair \(i, j \in M\), there are edges \(e_{k i j}\) and \(e_{k j i} \in E\) for all vehicles \(k \in V\).

Under this setting, a vehicle \(v_k\) should start at \(s_k\), follow the edges to visit a subset of task nodes (often together with other vehicles), and finish at \(u_k\). Based on graph \(G\), an integer program can be formulated. Here we define common notations, where the decision variables are \{\(x\), \(y\), \(z\), \(q\)\}.

\hskip-0.22in
\begin{tabular}{cp{0.06\linewidth}p{0.85\linewidth}} 
    &\(x_{k i j}\)& =1, if vehicle \(k \in V\) follows edge \(e_{k i j}\), where \(i \in M \cup S\) and \(j \in M \cup U\), otherwise 0.
    \\
    &\(y_{k i}\)& =1, if task \(i \in M\) is visited by vehicle \(k \in V\), otherwise 0.
    \\
    &\(z_{i}\)& =1, if task \(i \in M\) is completed, otherwise 0.
    \\
    &\(q_i\)& Time that task \(i \in M\) begins.
    \\
    &\(b_{k i j}\)& Stochastic energy cost to travel edge \(e_{k i j}\) (\(k \in V, i \in M \cup S, j \in M \cup U\)). Assume Gaussian distributed, i.e. \(b_{k i j} \sim N(\mu_{k i j}, \sigma_{k i j})\). More details are in Sec. \ref{sec:uncertainty_model}.
    \\
    &\(t_{k i j}\)& Deterministic time needed to travel edge \(e_{k i j}\) (\(k \in V, i \in M \cup S, j \in M \cup U\)).
    \\
    &\(t_{k i}\)& Deterministic time for vehicle \(k \in V\) to complete its part for task \(i \in M\).
    \\
    &\(C_{q}\)& Time penalty coefficient.
    \\
    &\(C_{\text{large}}\)& Large constant number for integer programming.
    \\
    &\(B_k\)& Energy capacity of vehicle \(k \in V\).
\end{tabular}
\vspace{0.05\in}

Now we formally define the deterministic model.

\subsubsection{Objective Function}
The objective function (\ref{eqn:deterministic_objective}) minimizes the expected sum of the energy cost of the whole vehicle fleet and the sum of the times that vehicles arrive at the terminal nodes. \(C_q\) is set to a small number such that energy is the primary optimization objective.
\begin{align}
    \min & \quad E ( \underset{k \in V}{\sum} \ \underset{i \in M \cup S}{\sum} \ \underset{j \in M \cup U}{\sum} b_{k i j} \cdot x_{k i j} ) + C_{q} \underset{i \in U}{\sum} q_i \nonumber \\
    &= \underset{k \in V}{\sum} \ \underset{i \in M \cup S}{\sum} \ \underset{j \in M \cup U}{\sum} \mu_{k i j} \cdot x_{k i j} + C_{q} \underset{i \in U}{\sum} q_i \label{eqn:deterministic_objective}
\end{align}

\subsubsection{Basic Range Constraints}
Constraints (\ref{eqn:integer_constraint}) and (\ref{eqn:time_constraint2}) define the nature of the variables.
\begin{align}
    x_{kij}, y_{ki}, z_{i} &\in \{0,1\} \quad \forall i,j \in S \cup M \cup U, \forall k \in V \label{eqn:integer_constraint} \\
    q_i \geq 0 \quad &\forall i \in M \cup U, \quad q_i = 0 \quad \forall i \in S \label{eqn:time_constraint2} 
\end{align}

\subsubsection{Network Flow Constraint}
Constraints (\ref{eqn:flow_constraint1}), (\ref{eqn:flow_constraint2}), and (\ref{eqn:flow_constraint3}) are network flow constraints, which ensure that a vehicle follows a route from the start to the terminal. (\ref{eqn:node_visited_constraint}) reflects the relation that a task is visited by a vehicle if there is a corresponding incoming edge selected.
\begin{align}
    \underset{i \in S \cup M} {\sum} x_{k i m} &= \underset{j \in U \cup M} {\sum} x_{k m j} \quad \forall m \in M, \forall k \in V \label{eqn:flow_constraint1} \\
    \underset{i \in M} {\sum} x_{k s i} &\leq 1 \quad \forall k \in V \label{eqn:flow_constraint2} \\
    y_{k j} &\leq \underset{i \in M} {\sum} x_{k s i} \quad \forall k \in V \label{eqn:flow_constraint3} \\
    y_{k j} &= \underset{i \in S \cup M} {\sum} x_{k i j} \quad \forall j \in M, \forall k \in V \label{eqn:node_visited_constraint}
\end{align}

\subsubsection{Energy Constraints}
The total energy cost for a vehicle should not exceed its energy capacity \(B_k\). In a deterministic model, this energy capacity constraint is imposed by limiting the expected total energy cost as in (\ref{eqn:energy_constraint}).
\begin{align}
    \underset{i \in M \cup S}{\sum} \ & \underset{j \in M \cup U}{\sum} \mu_{k i j} \cdot x_{k i j} \leq B_k \quad \forall k \in V \label{eqn:energy_constraint}
\end{align}

\subsubsection{Time Constraints}
(\ref{eqn:time_constraint1}) is a time scheduling constraint: for a given vehicle, the difference between the start time of the next task and the current task should be larger than the service time at the current task plus the travel time. This ensures that all the vehicles in a team are able to arrive before a new task starts. This constraint also eliminates subtours.
\begin{align}
    q_{i} - q_{j} &+ t_{k i j} + t_{k i} \leq C_{\text{large}} (1 - x_{k i j}) \nonumber\\
    &\forall i \in S \cup M, \ \forall j \in U \cup M, \ \forall k \in V \label{eqn:time_constraint1}
\end{align}

\subsubsection{Task Constraints}
Constraint (\ref{eqn:task_complete_constraint2}) ensures that all the tasks are completed. (\ref{eqn:task_complete_constraint1}) reflects the requirement of a task: \(i \in M\) can be completed only if the ``sum capabilities" of the vehicle team drive the task requirement function \(\rho_i(\cdot)\) to 1. This is an innovative constraint that encodes the vehicle and task heterogeneity. As the capability set can be defined flexibly, this enables resilient teaming for a broad set of practical problems. 
\begin{align}
    &z_{i} = 1 \quad \forall i \in M \label{eqn:task_complete_constraint2} \\
    &z_{i} \leq \rho_i(\underset{k \in V}{\sum} c_{k 1} \cdot y_{k i}, \ \underset{k \in V}{\sum} c_{k 2} \cdot y_{k i}, \ \cdots) \quad \forall i \in M  \label{eqn:task_complete_constraint1}
\end{align}

\subsubsection{Constraints Linearization}
All the components in the above formulation are linear except for \(\rho_i(\cdot)\) in (\ref{eqn:task_complete_constraint1}). Now we take (\ref{eqn:task_requirement_function}) as an example to show how to represent constraint (\ref{eqn:task_complete_constraint1}) using linear constraints. The input to \(\rho_{m_1}(\cdot)\) in (\ref{eqn:task_requirement_function}) and  (\ref{eqn:task_complete_constraint1}) should be equal; therefore, we have (\ref{eqn:alpha_equation}). The linear representation for (\ref{eqn:task_complete_constraint1}) (when \(i=m_1\) and \(\rho_{m_1}\) as in (\ref{eqn:task_requirement_function})) is (\ref{eqn:linear_equivalent}). With this, the deterministic formulation becomes a mixed integer linear program (MILP).
\begin{align}
    & \alpha_a = \underset{k \in V}{\sum} c_{k a} \cdot y_{k 1} \quad \forall a \in A \label{eqn:alpha_equation}\\
    & w_a \leq \alpha_a / \gamma_a, \quad w_a \geq (\alpha_a - \gamma_a + 1) / C_\text{large} \quad \forall a \in A \nonumber \\
    & z_{m_1} \leq w_{a_2} + w_{a_3}, \ \ \
    z_{m_1} \leq w_{a_4}, \ \ \
    z_{m_1} \leq w_{a_5} + w_{a_6} \nonumber \\
    & w_a \in \{0,1\} \quad \forall a \in A \label{eqn:linear_equivalent}
\end{align}

\subsection{Chance Constrained Programming Model}\label{sec:ccp_model}

The deterministic model uses the expected energy cost \(\mu_{kij}\) in the place of \(b_{kij}\) to pose an energy capacity constraint. This is problematic when \(b_{kij}\) has a wide distribution, placing the vehicles at a high risk of energy failure. To address this, a chance constrained programming ({\color{black}CCP}) model and stochastic programming model with recourse ({\color{black}SPR}) are proposed.

One issue with constraint (\ref{eqn:energy_constraint}) is that it does not provide a probabilistic bound for whether an energy failure will happen. Here we change it to a chance constraint and require the probability that vehicle \(k \in V\) does not reach the energy limit to be larger than a confidence level \(\beta_k\), i.e. replace constraint (\ref{eqn:energy_constraint}) with (\ref{eqn:chance_constraint}). Therefore, the CCP model is defined using (\ref{eqn:deterministic_objective})-(\ref{eqn:node_visited_constraint}), (\ref{eqn:time_constraint1})-(\ref{eqn:task_complete_constraint1}), and (\ref{eqn:chance_constraint}).
\begin{align}
    P(\underset{i \in M \cup S}{\sum} \ \underset{j \in M \cup U}{\sum} b_{k i j} \cdot x_{k i j} \leq B_k ) \geq \beta_k \quad \forall k \in V \label{eqn:chance_constraint}
\end{align}
\begin{align}
    \psi^{-1}(\beta_k) &\sqrt{ \underset{i \in M \cup S}{\sum} \ \underset{j \in M \cup U}{\sum} \sigma_{k i j}^2 \cdot x_{k i j} } \nonumber \\
    &+ \underset{i \in M \cup S}{\sum} \ \underset{j \in M \cup U}{\sum} \mu_{k i j} \cdot x_{k i j}
    \leq
    B_k \quad \forall k \in V \label{eqn:deterministic_equivalent}
\end{align}

Assume the energy cost to traverse an edge is Gaussian distributed \(b_{k i j} \sim N(\mu_{k i j}, \sigma_{k i j})\) and independent with different \(k\), \(i\), or \(j\). Let the cumulative distribution function of a standard Gaussian distribution be \(\psi(\cdot)\). The deterministic equivalent of (\ref{eqn:chance_constraint}) is (\ref{eqn:deterministic_equivalent}). 
{\color{black} The derivation of the distribution parameters are in Sec. \ref{sec:uncertainty_model}.} Note that the chance constraint (\ref{eqn:chance_constraint}) will strongly affect the feasible domain of decision variables when the probability distribution of \(b_{kij}\) is wide.

\subsection{Stochastic Programming Model with Recourse}\label{sec:spr_model}
The CCP model has several limitations: first, if the variance of the probability distribution is large, this can result in over-conservatism or no feasible solution; second, the CCP model does not consider the possibility of route failures; third, it does not consider additional recourse costs when failures occur. As an alternative, an SPR model, which allows for failures and considers a recourse strategy beforehand, is able to address these limitations.

Based on the proposed problem, we consider two recourse actions when a vehicle runs out of energy during the mission. First, a rescue vehicle (e.g. a truck) will visit the failed vehicle and either refuel it or take it back to the start location. Second, another vehicle of the same type will move to the next task and take over all remaining tasks. Each of these actions introduces a recourse cost.

\begin{figure}[hbt!]
	\centering
	\includegraphics[width=0.83\linewidth]{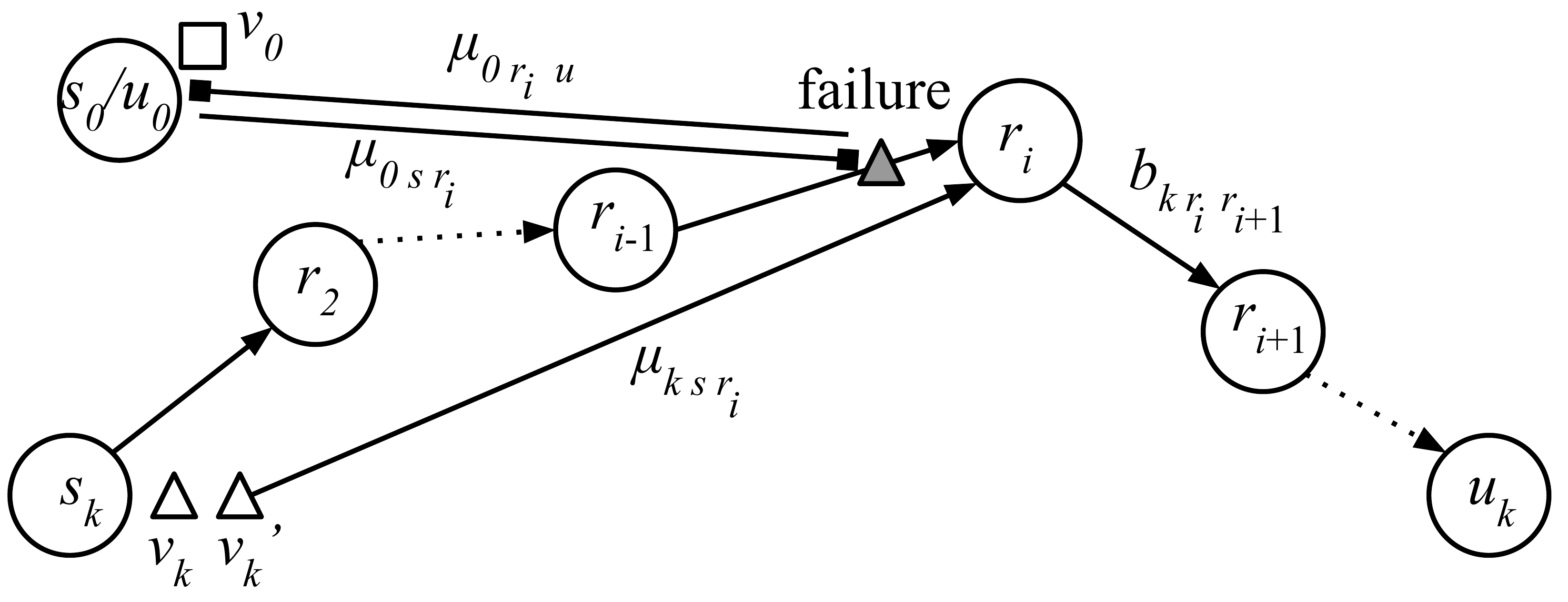}
	\caption{Recourse action. Dashed arrows stand for multiple edges and nodes in between. \(v_k'\) is of the same vehicle type as \(v_k\). \(v_0\) is a rescue truck, taking the failed vehicle back to the repair plant. Usually \(s_0\), \(u_0\), \(s_k\), and \(u_k\) are at the same location.}
	\label{fig:recourse_action}
\end{figure}

To demonstrate recourse costs, consider the following (Fig. \ref{fig:recourse_action}): given a routing plan, i.e. a solution to \(x_{kij}\), vehicle \(k \in V\) will follow route \(R_k = \{e_{k r_1 r_2}, e_{k r_2 r_3}, \cdots, e_{k r_{(n_r-1)} r_{n_r}}\}\), i.e. \(e_{k i j} \in R_k, \forall x_{k i j} = 1\). Note that \(r_1 = s_k\) and \(r_{n_r} = u_k\), indicating the start and terminal node for vehicle \(k\). In the event of vehicle \(k\) failing during edge \(e_{k r_{(i-1)} r_i}\), we assume the recourse costs to be \(\mu_{0 s r_i} + \mu_{0 r_i u} + \mu_{k s r_i}\). This describes the expected energy cost for a rescue vehicle \(0\) to travel from its start \(s_0\) to node \(r_i\) and then to terminal \(u_0\), and another vehicle of the same type to travel from its start \(s_k\) to task \(r_i\). The probability that the \(l\)-th failure of vehicle \(k\) occurs on edge \(e_{k r_{(i-1)} r_i}\) is \(P( \stackrel{i-1}{ \underset{j = 2} {\sum} } b_{k r_{(j-1)} r_j} \leq l\cdot B_k \leq \stackrel{i}{ \underset{j = 2} {\sum} } b_{k r_{(j-1)} r_j} )\) \cite{laporte2002integer,mendoza2013multi}. Therefore, the expected recourse cost \(g_k\) is given in (\ref{eqn:recourse_function}), where \(C_g \geq 1\) is the recourse penalty coefficient.

The SPR model minimizes the sum of the original cost and the recourse cost (\ref{eqn:spr_objective}), subject to constraints (\ref{eqn:integer_constraint})-(\ref{eqn:task_complete_constraint1}). Note that the deterministic energy constraint (\ref{eqn:energy_constraint}) is preserved in the SPR model to balance the workload between vehicles. Without this constraint in the SPR model, some vehicles may be assigned to a long route and fail multiple times, while others are highly underutilized.
\begin{align}
    g_k = C_g & \stackrel{n_r}{ \underset{i = 2} {\sum}} \stackrel{i-1}{ \underset{l = 1} {\sum}} ( P( \stackrel{i-1}{ \underset{j = 2} {\sum} } b_{k r_{(j-1)} r_j} < l\cdot B_k \leq \stackrel{i}{ \underset{j = 2} {\sum} } b_{k r_{(j-1)} r_j} ) \nonumber \\
    &\cdot (\mu_{k s r_i} + \mu_{0 s r_i} + \mu_{0 r_i u}) ) \quad \forall k \in V \label{eqn:recourse_function}\\
    \min \underset{k \in V}{\sum} &  \underset{i \in M \cup S}{\sum} \underset{j \in M \cup U}{\sum} \mu_{k i j} \cdot x_{k i j} + C_{q} \underset{i \in U}{\sum} q_i + \underset{k \in V}{\sum} g_k \label{eqn:spr_objective}
\end{align}

\section{Optimization Algorithms}\label{sec:algorithm}
The deterministic model is a mixed integer linear program (MILP) and can be solved by a number of off-the-shelf commercial solvers through branch and bound algorithms. This section will focus on the presentation of the algorithms used to solve the CCP and SPR models.

\subsection{Branch and Cut Algorithm for CCP Model}

To solve the proposed CCP model, a branch and cut algorithm based on the one in \cite{chen2014optimizing} is used. Noticing that without the nonlinear chance constraint, the model becomes a MILP, it is solved with constraint  (\ref{eqn:deterministic_equivalent}) relaxed. The details are in Algorithm \ref{alg:ccp_algorithm}, where checkNonlinearConstraints() is a function that returns the set of violated constraints.

\begin{algorithm}
\small
  Define problem \(p\) with objective fcn (\ref{eqn:deterministic_objective}) and constraints (ignore (\ref{eqn:deterministic_equivalent}))
  
  \While{True}{
    {\color{ForestGreen}// Iteratively solve problem \(p\)}
    
    Solve problem \(p\) and let \(x^p_{kij}\) be the solution.

    \If{\(x^p_{kij} = \O\)}{
      \Break {\color{ForestGreen}\textnormal{// Return no solution}}
    }

    violations \(=\) checkNonlinearConstraints((\ref{eqn:deterministic_equivalent}))

    \eIf{\textnormal{violations} \(\neq \O\)}{
      Cut the solution by adding a linear feasibility cut (\ref{eqn:switch_solution_constraint}) to problem \(p\) {\color{ForestGreen}{// Solve it again}}
    }
    {
      \Break {\color{ForestGreen}\textnormal{// Return the solution}}
    }
  }
  
  \caption{An Algorithm for the CCP Model}
  \label{alg:ccp_algorithm}
\end{algorithm}

In (\ref{eqn:switch_solution_constraint}), \(V_e\) is the set of vehicles that violate constraint (\ref{eqn:deterministic_equivalent}). Otherwise, if there is no violation, we obtain an optimal solution for the CCP problem, and stop the algorithm.
\begin{align}
    \underset{(i, j) \in R^p_k}{\sum} (1 - x_{k i j}) \geq 1 \quad \forall k \in V_e, \label{eqn:switch_solution_constraint} 
\end{align}
where $R^p_k = \ \{(i,j) \ | \ i,j \in  M$ and $x^p_{kij} = 1\}.$ 


\subsection{Integer L-shaped Algorithm for SPR Model}

To solve the proposed SPR model, we propose an integer L-shaped algorithm similar to the one in \cite{laporte2002integer}, 
where a branch and cut algorithm operates on a relaxed problem in each iteration until completing a tree-structure problem list \(\mathcal{P}\). Let the linear part (the first two sums) in (\ref{eqn:spr_objective}) be \(f\), and let the nonlinear part (the third sum) be \(g\). For each \(g_k \ (k \in V)\), we add a new variable \(\theta_k\) to represent its lower bound. Therefore, \(\theta = \sum \theta_k\) is the lower bound for \(g = \sum g_k\). For a problem in the list, the objective function (\ref{eqn:spr_relaxed_objective}) is optimized instead of (\ref{eqn:spr_objective}), i.e. \(f+\theta\) is optimized instead of \(f+g\). During the iterations, the lower bound \(\theta\) is tightened until \(\theta=g\) and an optimal solution is found. The details are in Algorithm \ref{alg:spr_algorithm}, where checkConstraints() and checkIntegerConstraints() are functions that return the set of violated constraints.

\newif\ifspralgorithm
\spralgorithmtrue
\ifspralgorithm
\begin{algorithm}
\small
  Define problem \(p\) with objective fcn (\ref{eqn:spr_relaxed_objective}) and no constraints
  
  \(f^* = +\infty\), \(g^* = +\infty\)
  
  \While{\(p \neq \O\)}{
    \While{True}{
      {\color{ForestGreen}// Iteratively solve problem \(p\)}
    
      Solve \(p\) and let \((x^p_{kij}, f^p, \theta^p)\) be the solution

      \If{\(x^p_{kij} = \O\)}{
        \Break {\color{ForestGreen}\textnormal{// No solution for problem \(p\)}}
      }

      \If{\(f^p+\theta^p \geq f^*+g^*\)}{
        {\color{ForestGreen}// The relaxed problem has a worse objective than the current best solution}
        
        \Break {\color{ForestGreen}\textnormal{// No need to explore \(p\) more}}
      }
      violations1 \(=\) checkConstraints((\ref{eqn:time_constraint2})-(\ref{eqn:task_complete_constraint1}))
      
      \If{\textnormal{violations1} \(\neq \O\)}{
        Add one violated constraint to problem \(p\)
        
        \Continue {\color{ForestGreen}\textnormal{// Solve it again}}
      }
      violations2 \(=\) checkIntegerConstraints((\ref{eqn:integer_constraint}))

      \If{\textnormal{violations2} \(\neq \O\)}{
        Branch problem \(p\) on a fractional variable into \(p_1\) and \(p_2\)
        
        Add both subproblems \(p_1\) and \(p_2\) into list \(\mathcal{P}\)
        
        \Break {\color{ForestGreen}\textnormal{// Start to solve \(p_2\) }}
      }
      {\color{ForestGreen}// Now a feasible solution to the linear relaxation}
      
      Compute \(g^p\) based on \((x^p_{kij}, f^p, \theta^p)\)
      
      \If{\(f^p + g^p < f^* + g^*\)}{
        \(f^* = f^p, g^* = g^p\)
      }
      
      \eIf{\(\theta^p \geq g^p\)}{
        {\color{ForestGreen}{// A nonlinearly optimal solution to problem \(p\) }}
        
        \Break
      }
      {
        Tighten the linear relaxation by adding an optimality cut (\ref{eqn:recourse_cut}) to problem \(p\)
        
        \Continue
      }
    }
    \(p = \) pop the last problem in list \(\mathcal{P}\)
  }
  \caption{\small An L-shaped Algorithm for the SPR Model}
  \label{alg:spr_algorithm}
\end{algorithm}
In line 25, we conclude that if \(\theta^p \geq g^p\), then the current solution minimizes the nonlinear objective function (\ref{eqn:spr_objective}) given the feasible region defined by the constraints of problem \(p\). 
To prove this, let another feasible point and the corresponding objective values be \((x^a_{kij}, f^a, g^a, \theta^a)\). As the current solution is linearly optimal for problem \(p\) and \(\theta^a\) is a lower bound for \(g^a\), we have \(f^p+\theta^p \leq f^a+\theta^a \leq f^a+g^a\). Recall that \(\theta^p \geq g^p\), and therefore, \(f^p+g^p \leq f^a+g^a\).

Otherwise, if \(\theta^p < g^p\), we tighten the relaxation at line 29 by adding the optimality cut (\ref{eqn:recourse_cut}). The cut means, \(\theta_k\) is tightened to \(g_k^p\) if we obtain a solution \(x_{kij}\) where the current route or a route containing the current route is traveled. \(n^p_k\) is the number of \((i,j)\) pairs in the set \(R^p_k\), i.e., the current route.
\else
\noindent Step 0: Initialize a basic problem \(p\) with the objective function (\ref{eqn:spr_relaxed_objective}) and no constraints, and add it to the problem list \(\mathcal{P}\). Let the optimal objective value \(f^* = +\infty\) and \(g^* = +\infty\).
\\
Step 1: Select a problem \(p\) from the list \(\mathcal{P}\). If none, break. \\
Step 2: Solve the the problem \(p\) and let \((x^p_{kij}, f^p, \theta^p)\) be the solution. If \(f^p+\theta^p \geq f^*+g^*\), i.e. the relaxed problem has a worse objective than the current best solution, complete \(p\) and go to Step 1.
\\
Step 3: Check constraints (\ref{eqn:flow_constraint1})-(\ref{eqn:energy_constraint}). If there are violations, add one violated constraint to \(p\), and go to Step 2.
\\
Step 4: Check integer constraints (\ref{eqn:integer_constraint}). If there are violations, branch on a fractional variable and add both subproblems to \(\mathcal{P}\), drop \(p\), and go to Step 1.
\\
Step 5: Now we have a feasible solution for the original problem, therefore, we can compute \(g^p\) according to (\ref{eqn:recourse_function}). If \(f^p+g^p < f^*+g^*\), let \(f^* = f^p, g^* = g^p\).
\\
Step 6: If \(\theta^p \geq g^p\), then complete problem \(p\) and go to Step 1. This is because there is no possibility that another feasible point of problem \(p\) is with a lower objective value. To prove this, let another feasible point and the corresponding objective values be \((x^a_{kij}, f^a, g^a, \theta^a)\). As the current solution is linearly optimal for problem \(p\), \(f^p+\theta^p \leq f^a+\theta^a \leq f^a+g^a\). Recall that \(\theta^p \geq g^p\), and therefore, \(f^p+g^p \leq f^a+g^a\).
\\
Otherwise, if \(\theta^p < g^p\), we tighten the relaxation by adding the optimality cut (\ref{eqn:recourse_cut}). Then, go to Step 2. The cut means, \(\theta_k\) is tightened to \(g_k^p\) if we obtain a solution \(x_{kij}\) where the current path or a path containing the current path is traveled. \(n^p_k\) is the number of \((i,j)\) pairs in the set \(R^p_k\) (i.e. the current path).
\fi

If the algorithm terminates with \(f^* \neq +\infty\), then the solution associated with \(f^*\) and \(g^*\) is the optimal solution, otherwise, there is no feasible solution for the SPR model.
\begin{align}
    \min \underset{k \in V}{\sum} & \underset{i \in M \cup S}{\sum} \underset{j \in M \cup U}{\sum} \mu_{k i j} \cdot x_{k i j} + C_{q} \underset{i \in U}{\sum} q_i + \underset{k \in V}{\sum} \theta_k \label{eqn:spr_relaxed_objective}
\end{align}
\begin{align}
    \theta_k &\geq g^p_k \cdot ( - n^p_k + 1 + \underset{(i, j) \in R^p_k}{\sum} x_{kij} ) \quad \forall k \in V, \label{eqn:recourse_cut}
\end{align}
where $R^p_k = \{(i,j) \ | \ i,j \in  M \cup S \cup U$ and $x^p_{kij} = 1\}.$

\section{Uncertainty Model}\label{sec:uncertainty_model}

\newif\ifremovegpr
\removegprtrue
\ifremovegpr

For the optimization models in Sec. \ref{sec:optimization_model}, parameters \(C_q\) and \(C_{\text{large}}\) are manually defined. \(c_{ka}\) and \(D_k\) can be obtained according to vehicle properties. \(t_{ki}\) can be estimated according to the task requirements. As we are considering off-road operations, we assume that \(t_{kij}\) is proportional to the path length between node \(i\) and \(j\). In off-road operations, the edge energy cost, \(b_{kij}\), is uncertain and needs further modeling.

In environments where the energy cost to traverse a unit length varies with respect to the position and is uncertain, the energy field can be modeled as a Gaussian process \cite{quann2017energy,quann2018ground,quann2019chance}. A Gaussian process is a collection of random variables with joint Gaussian distribution \cite{williams2006gaussian}. Based on a set of samples of the unit energy cost from sensor data (either collected online or from previous explorations), a Gaussian process map similar to Fig. \ref{fig:energy_map_and_path} can be predicted.

\begin{figure}[hbt!]
	\centering
    \includegraphics[width=0.55\linewidth, trim=0 0 0 0, clip]{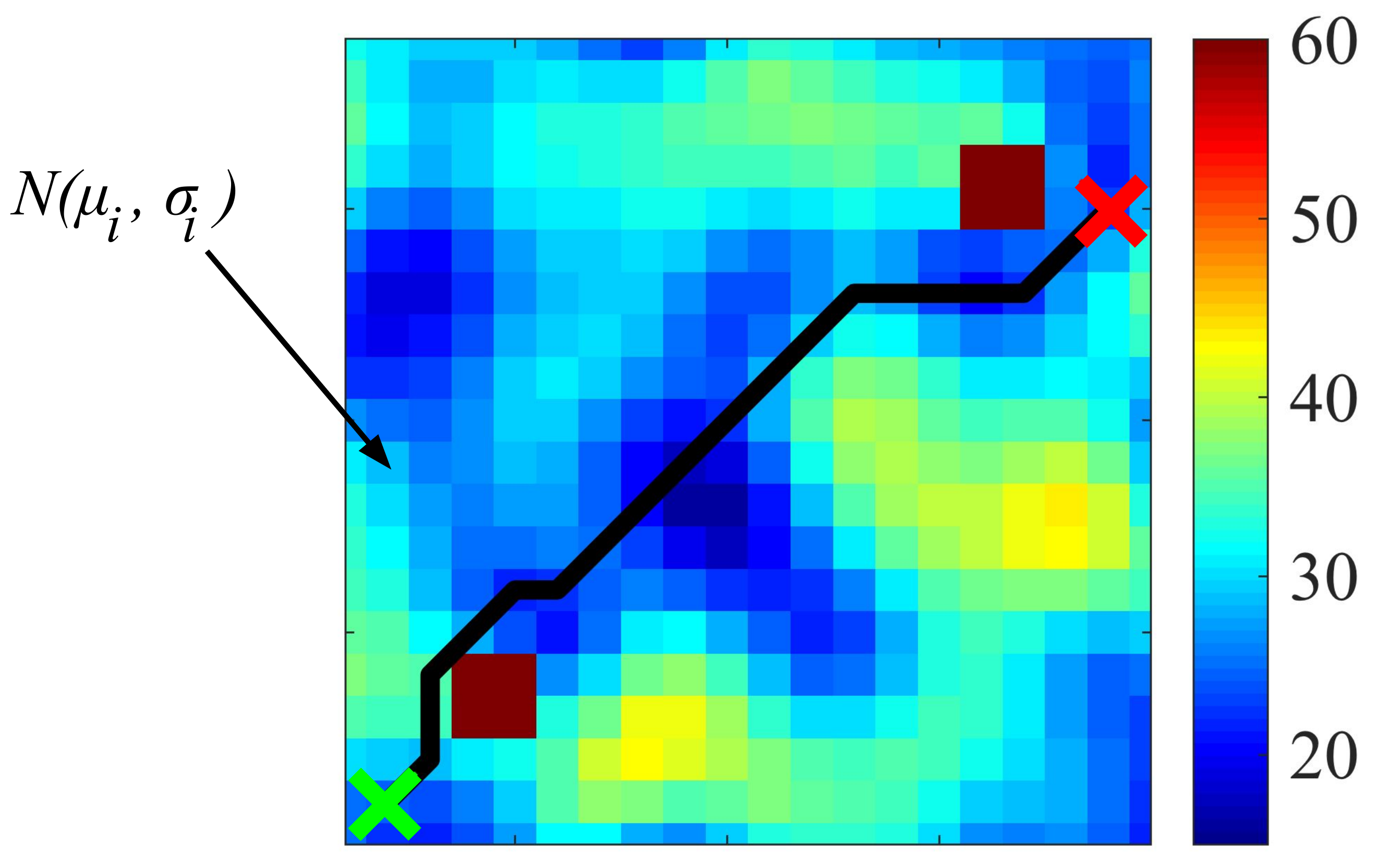}
	\caption{An example of a Gaussian process energy map. Each cell is a Gaussian variable that encodes the energy cost to travel a unit length. The red cells are obstacles. The black line is the minimum expected energy cost path between the green and red x's.}
	\label{fig:energy_map_and_path}
	\squeezeup
    \vspace{-2mm}
\end{figure}

Based on the energy map, we can use the A* algorithm \cite{pohl1973avoidance} to calculate the minimum expected energy cost path \cite{quann2019chance} between task locations while avoiding obstacles. For example, if two tasks are at the green and red x's in Fig. \ref{fig:energy_map_and_path}, we can evaluate the uncertain energy cost between the tasks by computing the black path and then the path cost.

Suppose the computed path contains ordered cell indices \(\mathcal{L} = \{l_1,\cdots,l_{n_\mathcal{L}}\}\). Let the distance between two cells \(i\) and \(j\) be \(d_{ij}\). Suppose the unit energy cost at cell \(i\) follows distribution \(N(\mu_i, \sigma_i)\). Define \(\mathbf{d}_\mathcal{L} = [ d_{l_1 l_2},\cdots,d_{l_{n_\mathcal{L}-1} l_{n_\mathcal{L}}}]\transpose\). Define \(\mathbf{m}_\mathcal{L} = [\mu_{1}, \cdots, \mu_{n_\mathcal{L}-1}]\transpose\), and let the corresponding covariance matrix be \(\mathbf{\Sigma}_\mathcal{L}\). Then, for a specific vehicle, the stochastic cost to travel between two tasks, \(b_\mathcal{L}\), is a Gaussian distribution, i.e. \(b_\mathcal{L} \sim \mathcal{N}(\mu_\mathcal{L},\sigma_\mathcal{L})\), and can be calculated according to (\ref{eqn:path_mu}) and (\ref{eqn:path_sigma}).
\begin{align}
    \mu_\mathcal{L} &= \stackrel{n_\mathcal{L}-1}{ \underset{i=1}{\sum}}{\mu_{l_i} \cdot d_{l_i l_{i+1}}} = \mathbf{d}_\mathcal{L}\transpose \mathbf{m}_\mathcal{L} \label{eqn:path_mu}\\
    \sigma_\mathcal{L} &= \mathbf{d}_\mathcal{L}\transpose \mathbf{\Sigma}_\mathcal{L} \mathbf{d}_\mathcal{L} \label{eqn:path_sigma}
\end{align}

\else
For the optimization models in Sec. \ref{sec:optimization_model}, parameters \(C_q\) and \(C_{\text{large}}\) are manually defined. \(c_{ka}\) and \(D_k\) can be obtained according to vehicle properties. \(t_{ki}\) can be estimated according to the task requirements. As we are considering off-road operations, we assume that \(t_{kij}\) is proportional to the path length between node \(i\) and \(j\). In off-road operations, the edge energy cost, \(b_{kij}\), is uncertain and needs further modeling. In this section, we show how \(b_{kij}\) is modeled as a Gaussian variable and obtained from sensing data.

\subsection{Energy Map}\label{sec:energy_map}

In off-road or other environments where the energy cost to traverse a unit length varies with respect to the position and is uncertain, the energy field can be modeled as a Gaussian process \(\{h^p_i,\mathbf{p}^p_i\} \ (i = 1, \cdots,n_p)\) \cite{quann2017energy}, \cite{quann2018ground}, \cite{quann2019chance}, where \(\mathbf{p}^p_i\) is the \(i\)-th location in the map and \(h^p_i\) is the corresponding energy cost for traversing a unit distance. A Gaussian process is a collection of random variables with joint Gaussian distribution \cite{williams2006gaussian}.

Suppose we are given a set of sensing data \(\{h^s_i,\mathbf{p}^s_i\} \ (i = 1, \cdots,n_s)\), where \(\mathbf{p}^s_i\) is the \(i\)-th location and \(h^s_i\) is the corresponding sensing data of the energy cost. Let \(\mathbf{h}_s = [h^s_1,h^s_2,\cdots,h^s_{n_s}]\transpose\), \(\mathbf{h}_p = [h^p_1,h^p_2,\cdots,h^p_{n_p}]\transpose\), then \(\mathbf{h}_s\) and \(\mathbf{h}_p\) are jointly Gaussian distributed (\ref{eqn:h_gaussian}), where \(\mu\) and \(\mathbf{\Sigma}\) are the mean and covariance matrix. \(\mathbf{\mu}_s\), \(\mathbf{\mu}_{p}\), \(\mathbf{\Sigma}_{s}\), \(\mathbf{\Sigma}_{sp}\),  \(\mathbf{\Sigma}_{ps}\), and \(\mathbf{\Sigma}_{p}\) are the matrix blocks.
\begin{align}
    \left[ \begin{matrix*}[l]
    \mathbf{h}_s \\ \mathbf{h}_p
    \end{matrix*} \right]
    \sim
    \mathcal{N}
    \left(
    \mathbf{\mu}, \mathbf{\Sigma}
    \right)
    =
    \mathcal{N}
    \left(
    \left[ \begin{matrix*}[l]
    \mathbf{\mu}_s \\ \mathbf{\mu}_p
    \end{matrix*} \right]
    ,
    \left[ \begin{matrix*}[l]
    \mathbf{\Sigma}_s + \sigma_s^2 \mathbf{I} & \mathbf{\Sigma}_{sp} \\
    \mathbf{\Sigma}_{ps}                   & \mathbf{\Sigma}_{p}
    \end{matrix*} \right]
    \right)
    \label{eqn:h_gaussian}
\end{align}

It is assumed the means and covariances are known. Let the \(i\)-th entry in \(\mathbf{\mu}\) be \(\mu_i\), and the corresponding position be \(\mathbf{p}_i\). Let the \(ij\)-th element in \(\mathbf{\Sigma}\) be \(\Sigma_{ij}\). Then the means and covariances are calculated by a mean function \(m_h(\cdot)\) (set to a constant \(C_\mu\)) and a kernel function \(k_{h}(\cdot,\cdot)\) respectively in (\ref{eqn:mean_function})-(\ref{eqn:kernal_function2}). \(k_{h_1}(\cdot,\cdot)\) is the squared exponential (SE) kernel. \(k_{h_2}(\cdot,\cdot)\) is the white kernel, and is added because there is noise in the sensing data. \(||\cdot||\) is the Euclidean norm. \(\sigma_f\), \(\sigma_s\) are the standard deviation scaling factors and \(l_f\) is the length scale. These hyper-parameters need to be determined appropriately according to prior knowledge or an optimization based on the training data \cite{quann2017energy}. With this, the conditional distribution of the map entries \(\mathbf{h}_p\) given the sensor data \(\mathbf{h}_s\) can be calculated as (\ref{eqn:conditional1})-(\ref{eqn:conditional3}).
\begin{align}
    \mu_i &= m_h(\mathbf{p}_i) = C_\mu \label{eqn:mean_function}\\
    \Sigma_{ij} &= k_{h}(\mathbf{p}_i, \mathbf{p}_j) = k_{h_1}(\mathbf{p}_i, \mathbf{p}_j) + k_{h_2}(\mathbf{p}_i, \mathbf{p}_j) \label{eqn:kernal_function}
\end{align}
\vspace{-0.7cm}
\begin{align}
    k_{h_1}(\mathbf{p}_i, \mathbf{p}_j) &=
    \sigma_f^2 \exp(-\frac{1}{2 l_f^2}||\mathbf{p}_i - \mathbf{p}_j||^2) \label{eqn:kernal_function}
\end{align}

\begin{align}
    k_{h_2}(\mathbf{p}_i, \mathbf{p}_j) &= 
    \begin{cases}
    \sigma_s^2, \text{if } \mathbf{p}_i = \mathbf{p}_j \text{ and } \mathbf{p}_i \in \{\mathbf{p}^p_1,...\} \\
    0, \text{ otherwise}
    \end{cases} \label{eqn:kernal_function2} 
\end{align}
\vspace{-0.3cm}
\begin{align}
    \mathbf{h}_{p|s} &\sim \mathcal{N} (\mathbf{\mu}_{p|s}, \mathbf{\Sigma}_{p|s}) \label{eqn:conditional1} \\
    \mathbf{\mu}_{p|s} &= \mathbf{\mu}_{p} + \mathbf{\Sigma}_{p s} (\mathbf{\Sigma}_{s} + \sigma_s^2 \mathbf{I})^{-1} (\mathbf{h}_s - \mathbf{\mu}_s) \label{eqn:conditional2}\\
    \mathbf{\Sigma}_{p|s} &= \mathbf{\Sigma}_{p} - \mathbf{\Sigma}_{p s} (\mathbf{\Sigma}_{s} + \sigma_s^2 \mathbf{I})^{-1} \mathbf{\Sigma}_{s p}\label{eqn:conditional3}
\end{align}

\subsection{Edge Energy Cost and Uncertainty}

We use the method in Sec. \ref{sec:energy_map} to build an energy map. In the meantime, assume that the occupancy grid map of the obstacles in the environment is given, as in practice, there are many well-developed simultaneous localization and mapping systems that can output occupancy grid maps with high accuracy. With the energy map and occupancy map, we use the A* algorithm \cite{pohl1973avoidance} to calculate the minimum expected energy cost path \cite{quann2019chance} between task locations while avoiding the obstacles. Suppose the computed path contains ordered cell indices \(\mathcal{L} = \{l_1,\cdots,l_{n_\mathcal{L}}\}\). Let the distance between two cells \(i\) and \(j\) be \(d_{ij} = ||\mathbf{p}_i - \mathbf{p}_j||\). Let \(\mu^{p|s}_{i}\) be the \(i\)-th element in \(\mu_{p|s}\). Define \(\mathbf{d}_\mathcal{L} = [ d_{l_1 l_2},\cdots,d_{l_{n_\mathcal{L}-1} l_{n_\mathcal{L}}}]\transpose\). Define \(\mathbf{m}_\mathcal{L} = [\mu^{p|s}_{1}, \cdots, \mu^{p|s}_{n_\mathcal{L}-1}]\transpose\), and let the corresponding covariance matrix be \(\mathbf{\Sigma}_\mathcal{L}\). Then, for a specific vehicle, the stochastic cost to travel between two tasks \(b_\mathcal{L}\) is a Gaussian distribution, i.e. \(b_\mathcal{L} \sim \mathcal{N}(\mu_\mathcal{L},\sigma_\mathcal{L})\), and can be calculated according to (\ref{eqn:path_mu}) and (\ref{eqn:path_sigma}).
\begin{align}
    \mu_\mathcal{L} &= \stackrel{n_\mathcal{L}-1}{ \underset{i=1}{\sum}}{\mu^{p|s}_{l_i} \cdot d_{l_i l_{i+1}}} = \mathbf{d}_\mathcal{L}\transpose \mathbf{m}_\mathcal{L} \label{eqn:path_mu}\\
    \sigma_\mathcal{L} &= \mathbf{d}_\mathcal{L}\transpose \mathbf{\Sigma}_\mathcal{L} \mathbf{d}_\mathcal{L} \label{eqn:path_sigma}
\end{align}

\fi

\section{Experiments and Discussion}\label{sec:result_and_discussion}
In this section, we evaluate the models and algorithms proposed in the previous sections. First, a series of computational experiments are conducted to show the performance of the deterministic, CCP, and SPR models with respect to various vehicle and task numbers, vehicle and task types, and energy uncertainty levels. Second, a practical problem is solved to show one possible application of the models.

The models and algorithms are implemented using GUROBI with the C\(++\) interface. The experiments were done on a laptop with Intel i7-7660U CPU (2.50GHz).

\subsection{Computational Experiments}

\subsubsection{Setup and Test Cases}
The computational test focuses on the optimization models and algorithms; therefore, we assume the energy cost distributions of all edges are given. Suppose there are \(n_v\) vehicles, \(n_m\) tasks, \(n_a\) types of capabilities, \(n_{av}\) types of vehicles, and \(n_{am}\) types of tasks. The tasks are distributed randomly in the region \([0,640]\times[0,480]\), and the vehicles' start and terminal locations are identical and distributed randomly in the region \([310,330]\times[230,250]\).  Suppose the travel distance for edge \(e_e\) is \(d_e\), then the energy cost is assumed to be \(b_e \sim \mathcal{N}(C_\mu d_e, C_\sigma d_e)\) for the computational experiments, where \(C_\mu\) and \(C_\sigma\) are the hyper-parameters for the mean and standard deviation. By varying these hyper-parameters, we evaluate the performance of the three models and algorithms with respect to different problem sizes, uncertainty levels, and heterogeneity levels. The performance metrics used for evaluation include the value of the objective function and the computation time. For these computational experiments, \(C_\mu\) is fixed at \(30\) and the vehicle energy capacity, \(B_k\), is set to \(40000\) for all vehicles \(k \in V\) unless further explained. The confidence level, \(\beta_k\), in the CCP model is set to \(95\%\) for all vehicles. The recourse penalty coefficient, \(C_g\), in the SPR model is set to 1, which means the recourse costs are not weighted. The time penalty coefficient, \(C_q\), is set to 1 as well. Considering the order of magnitude for energy and time, this ensures that the energy cost is the primary optimization objective. When running the algorithms for the deterministic, CCP, and SPR models, the time limit is 500 seconds.

\subsubsection{Results and Discussion}
The objective value and computation time comparison are shown in TABLES \ref{tab:vehicle_number}-\ref{tab:uncertainty_level}, with \(f^*\) and \(g^*\) indicating the energy cost and recourse energy cost, respectively. Note that for each row in these tables, we randomly generated 5 test cases with different task and vehicle locations, and report the mean performances. According to the result, the model can be solved to optimal within a few seconds when \(n_v \leq 50\) and \(n_m \leq 12\). The computation time increases when the vehicle and task number increases. The algorithms are scalable to vehicle number, but not very scalable to task number. When \(n_v=6\), the largest \(n_m\) that the program can handle is around 30 for the three models (depending on the specific problems). When more heterogeneity is introduced to the problem, the computation time becomes smaller (TABLE \ref{tab:task_type_number}). This is because heterogeneity within the vehicles and tasks introduces further constraints and limits the search space of the algorithms.

The standard deviation (uncertainty level) has a direct impact on the CCP model, as it directly scales the \(\sigma_{kij}^2\) in (\ref{eqn:deterministic_equivalent}). When the uncertainty is large or \(\beta_k\) is set too high, the CCP model might return no solution. The uncertainty level has an implicit impact on the SPR model by influencing the expected recourse cost \(g^*\). According to TABLE \ref{tab:uncertainty_level}, both CCP and SPR result in larger objective values with larger \(C_\sigma\) values because they switch to more conservative solutions (shorter routes with lower possibility to fail).

\begin{table}[htb!]
  \centering
    \begin{tabular}{cccc@{\hskip -0.01\linewidth}ccc@{\hskip -0.01\linewidth}ccc}
    \toprule
    \multirow{2}[4]{*}{\(n_v\)} & \multicolumn{2}{c}{Deterministic} &       & \multicolumn{2}{c}{CCP} &       & \multicolumn{3}{c}{SPR} \\
\cmidrule{2-3}\cmidrule{5-6}\cmidrule{8-10}          & \(f^*\)  & Time  &       & \(f^*\)  & Time  &       & \(f^*\)  & \(f^*\)+ \(g^*\) & Time \\
    \midrule
    6     & 56215 & 0.1   &       & 57376 & 0.2   &       & 56726 & 57086 & 0.2 \\
    8     & 56199 & 0.2   &       & 57285 & 0.3   &       & 56627 & 56997 & 0.3 \\
    10    & 56161 & 0.3   &       & 57280 & 0.4   &       & 56597 & 56965 & 0.5 \\
    20    & 56093 & 0.9   &       & 56994 & 1.6   &       & 56566 & 56936 & 2.1 \\
    50    & 55875 & 5.1   &       & 56909 & 15.7  &       & 56344 & 56706 & 51.4 \\
    \bottomrule
    \end{tabular}%
  \caption{Objective value and computation time for different vehicle numbers \(n_v\). (\(n_m\)=6, \(n_a\)=2, \(n_{av}\)=2, \(n_{am}\)=3, \(C_\sigma\)=6.)} 
  \label{tab:vehicle_number}%
\end{table}%

\begin{table}[htb!]
\hskip-0.18cm
    \begin{tabular}{cccc@{\hskip -0.01\linewidth}ccc@{\hskip -0.01\linewidth}ccc}
    \toprule
    \multirow{2}[4]{*}{\(n_m\)} & \multicolumn{2}{c}{Deterministic} &       & \multicolumn{2}{c}{CCP} &       & \multicolumn{3}{c}{SPR} \\
\cmidrule{2-3}\cmidrule{5-6}\cmidrule{8-10}          & \(f^*\)  & Time  &       & \(f^*\)  & Time  &       & \(f^*\)  & \(f^*\)+ \(g^*\) & Time \\
    \midrule
    6     & 56215 & 0.1   &       & 57376 & 0.2   &       & 56726 & 57086 & 0.2 \\
    12    & 85139 & 4.6   &       & 85139 & 7.0   &       & 85142 & 85180 & 8.2 \\
    18    & 119078 & 41.2  &       & 121870 & 178.5 &       & 119078 & 119889 & 119.1 \\
    24    & \textcolor[rgb]{ .502,  .502,  .502}{125941} & \textcolor[rgb]{ .502,  .502,  .502}{305.6} & \textcolor[rgb]{ .502,  .502,  .502}{} & \textcolor[rgb]{ .502,  .502,  .502}{123316} & \textcolor[rgb]{ .502,  .502,  .502}{267.7} & \textcolor[rgb]{ .502,  .502,  .502}{} & \textcolor[rgb]{ .502,  .502,  .502}{126209} & \textcolor[rgb]{ .502,  .502,  .502}{126224} & \textcolor[rgb]{ .502,  .502,  .502}{307.8} \\
    30    & \textcolor[rgb]{ .502,  .502,  .502}{129041} & \textcolor[rgb]{ .502,  .502,  .502}{500.0} & \textcolor[rgb]{ .502,  .502,  .502}{} & \textcolor[rgb]{ .502,  .502,  .502}{128046} & \textcolor[rgb]{ .502,  .502,  .502}{500.0} & \textcolor[rgb]{ .502,  .502,  .502}{} & \textcolor[rgb]{ .502,  .502,  .502}{128793} & \textcolor[rgb]{ .502,  .502,  .502}{128793} & \textcolor[rgb]{ .502,  .502,  .502}{500.0} \\
    \bottomrule
    \end{tabular}%
  \caption{Objective value and computation time for different task numbers \(n_m\). (\(n_v\)=6, \(n_a\)=2, \(n_{av}\)=2, \(n_{am}\)=3, \(C_\sigma\)=6.) For \(n_m\geq24\), \(B_k=80000 \ \forall k \in V\). The gray text means these two test cases are not solved to optimal (time limit is not the only termination criterion). The optimality gaps here are around 10\%. This sub-optimality leads to the reversed order of \(f^*\), e.g., for \(n_m=30\) the order is CCP \(<\) SPR \(<\) Deterministic. }
  \label{tab:task_number}%
\end{table}%

\begin{table}[htb!]
\hskip-0.18cm
    \begin{tabular}{cccc@{\hskip -0.01\linewidth}ccc@{\hskip -0.01\linewidth}ccc}
    \toprule
    \multirow{2}[4]{*}{\(n_{am}\)} & \multicolumn{2}{c}{Deterministic} &       & \multicolumn{2}{c}{CCP} &       & \multicolumn{3}{c}{SPR} \\
\cmidrule{2-3}\cmidrule{5-6}\cmidrule{8-10}          & \(f^*\)  & Time  &       & \(f^*\)  & Time  &       & \(f^*\)  & \(f^*\)+ \(g^*\) & Time \\
    \midrule
    1     & 108029 & 19.1  &       & 109930 & 30.8  &       & 108056 & 108646 & 24.2 \\
    2     & 85638 & 0.3   &       & 86231 & 0.3   &       & 85641 & 85893 & 0.3 \\
    4     & 80014 & 0.3   &       & 81639 & 0.5   &       & 80331 & 80516 & 0.5 \\
    6     & 77106 & 0.1   &       & 78087 & 0.1   &       & 77273 & 77416 & 0.1 \\
    \bottomrule
    \end{tabular}%
  \caption{Objective value and computation time for different number of task types \(n_{am}\). (\(n_v\)=6, \(n_m\)=6, \(n_a\)=3, \(n_{av}\)=3, \(C_\sigma\)=6.)}
  \label{tab:task_type_number}%
\end{table}%

\begin{table}[htb!]
  \centering
    \begin{tabular}{cccc@{\hskip -0.01\linewidth}ccc@{\hskip -0.01\linewidth}ccc}
    \toprule
    \multirow{2}[4]{*}{\(C_\sigma\)} & \multicolumn{2}{c}{Deterministic} &       & \multicolumn{2}{c}{CCP} &       & \multicolumn{3}{c}{SPR} \\
\cmidrule{2-3}\cmidrule{5-6}\cmidrule{8-10}          & \(f^*\)  & Time  &       & \(f^*\)  & Time  &       & \(f^*\)  & \(f^*\)+ \(g^*\) & Time \\
    \midrule
    3     & 56215 & 0.1   &       & 56726 & 0.1   &       & 56726 & 56755 & 0.1 \\
    6     & 56215 & 0.1   &       & 57376 & 0.2   &       & 56726 & 57086 & 0.2 \\
    9     & 56215 & 0.2   &       & 58223 & 0.3   &       & 56888 & 57569 & 0.2 \\
    12    & 56215 & 0.1   &       & 58223 & 0.2   &       & 57611 & 57931 & 0.3 \\
    15    & 56215 & 0.1   &       & 59027 & 0.3   &       & 57611 & 58124 & 0.3 \\
    \bottomrule
    \end{tabular}%
  \caption{Objective value and computation time for different uncertainty \(C_\sigma\). (\(n_v\)=6, \(n_m\)=6, \(n_a\)=2, \(n_{av}\)=2, \(n_{am}\)=3.)}
  \label{tab:uncertainty_level}%
\vspace{-0.2cm}
\end{table}%

\vspace{-0.2cm}
Comparing the results of the CCP and SPR models in the four tables, the SPR model leads to lower costs than the CCP model even if the recourse cost \(g^*\) is considered. This shows the potential for over-conservatism with the CCP model.

\vspace{-0.2cm}
\subsection{Practical Problem Application}\label{sec:practical_problem_application}
\subsubsection{Setup and Test Case}
In this section, the CCP and SPR models are applied to a practical test case to evaluate the feasibility for practical applications. Consider an explore and breach mission in an area with forests and fields, consisting of multiple scout, breach, and material transport tasks. These tasks are independent of each other and there is no specific time requirement, except that a task can only be initiated when all the vehicles in the team arrive. The goal is to form vehicle teams to complete all of the tasks and finish the mission with the lowest expected travel energy cost. The capabilities, vehicles, and task sets are modeled as follows.

Capability set \(A = \{a_1, \cdots, a_8\}\). \(a_1\): scout capability; \(a_2\): move fast; \(a_3\): transport materials; \(a_4\): move quietly; \(a_5\): shoot smoke; \(a_6\): break barriers;  \(a_7\): clear mines; \(a_8\): armor.

Vehicle set \(V = \{v_1, v_2, \cdots\}\). In total, there are 18 vehicles distributed from 6 types of vehicles \(v_1\)-\(v_6\). The number of vehicles per type is chosen based on task requirements.
The following provides detailed descriptions for each vehicle; note that the brackets describe the actual reference vehicles.
\(v_1\): armed vehicle (high mobility multipurpose wheeled vehicle, HMMWV), a fast vehicle that can carry materials and has mounted guns.
\(v_2\): scout vehicle (MRZR), a fast, light, unarmored vehicle that can travel nearly silently.
\(v_3\): tank (M1A2 abrams), a large main tank with heavy armor.
\(v_4\): stryker (M1126 stryker), a combat vehicle with medium armor.
\(v_5\): earthmover (M9 combat earthmover), a vehicle that breaks barriers.
\(v_6\): minesweeper (M160 mine clearance system).
The vehicle capability vectors are defined accordingly in TABLE \ref{tab:vehicle_capability}.

\begin{table}[htbp]
  \centering
    \begin{tabular}{lcccccccc}
    \toprule
     & \(a_1\)     & \(a_2\)     & \(a_3\)     & \(a_4\)     & \(a_5\)     & \(a_6\)     & \(a_7\)     & \(a_8\) \\
    \midrule
    \(v_1\): Armed vehicle & 1     & 1     & 1     & 0     & 0     & 0     & 0     & 1 \\
    \(v_2\): Scout vehicle & 1     & 1     & 0     & 1     & 0     & 0     & 0     & 0 \\
    \(v_3\): Tank          & 0     & 0     & 0     & 0     & 0     & 0     & 0     & 20 \\
    \(v_4\): Stryker       & 0     & 0     & 0     & 0     & 1     & 0     & 0     & 5 \\
    \(v_5\): Earthmover & 0     & 0     & 0     & 0     & 0     & 1     & 0     & 0 \\
    \(v_6\): Minesweeper & 0     & 0     & 0     & 0     & 0     & 0     & 1     & 0 \\
    \bottomrule
    \end{tabular}%
  \caption{Vehicle capabilities: Each row is a \(\mathbf{c}_k\) vector, for \(k = v_1 \cdots v_6\).}
  \label{tab:vehicle_capability}%
\vspace{-0.3cm}
\end{table}%

Task set \(M = \{m_1, m_2, \cdots, m_{14}\}\). \(m_1\) scout: move towards the target location quickly and identify enemy forces.
\(m_2\) quiet scout: in addition to the requirements for \(m_1\), the vehicle team needs to move quietly in order to not be noticed by enemies.
\(m_3\) smoke cover: use smoke to cover breach points.
\(m_4\) lane creation: clear mines, remove barriers, and create secured lanes.
\(m_5\) breach: breach barriers and enemies.
\(m_6\) enemy push back: use heavy fire to push enemies back.
\(m_7\) materials transport.
The task \(m_8\)-\(m_{14}\) are the same as \(m_1\)-\(m_{7}\) respectively. The task requirement functions are defined in (\ref{eqn:practical_mission_requirement}). Note that if \(\gamma_{a} = 1\), then \(\alpha_{a} \geq \gamma_{a}\) is abbreviated to \(\alpha_{a}\), for all capabilities \(a \in A\).
\begin{align}
    \rho_{m_1}(\cdot) &= \rho_{m_8}(\cdot) = \alpha_{a_1} \land \alpha_{a_2}\nonumber \\
    \rho_{m_2}(\cdot) &= \rho_{m_9}(\cdot) = \alpha_{a_1} \land \alpha_{a_2} \land \alpha_{a_4} \nonumber \\
    \rho_{m_3}(\cdot) &= \rho_{m_{10}}(\cdot) = \alpha_{a_5}\nonumber \\
    \rho_{m_4}(\cdot) &= \rho_{m_{11}}(\cdot) = \alpha_{a_6} \land \alpha_{a_7} \land \alpha_{a_8} \label{eqn:practical_mission_requirement} \\
    \rho_{m_5}(\cdot) &= \rho_{m_{12}}(\cdot) = \alpha_{a_6} \land (\alpha_{a_8} \geq 10) \nonumber \\
    \rho_{m_6}(\cdot) &= \rho_{m_{13}}(\cdot) = (\alpha_{a_8} \geq 20) \nonumber \\
    \rho_{m_7}(\cdot) &= \rho_{m_{14}}(\cdot) = \alpha_{a_3} \nonumber
\end{align}

For the ground truth map, we randomly generate a map (size: \(100\times100\)) in region \([0,1000]^2\), with the cell values jointly Gaussian distributed. 
The underlying co-variance matrix is generated according to
\ifremovegpr
the Gaussian process method in \cite{quann2018ground} using a 2-dimensional squared exponential kernel.
\else
(\ref{eqn:mean_function}) and (\ref{eqn:kernal_function}), where (\(\sigma_f = 6, l_f = 100, \sigma_s = 0, C_\mu = 30\)). 
\fi
This map is generated to be the ground truth energy map for a 1-ton vehicle and {\color{black}scaled} by the weight of the 6 vehicles, i.e. [2.36, 0.879, 61.3, 19.0, 24.4, 10.0] tons. The vehicles' energy capacities are set to the following values proportional to their fuel tank size, \([B_{v_1}, \cdots, B_{v_6}]\) = [25.0, 7.25, 500, 160, 134, 80.0]\(\times 5700\). The ratio \(5700\) is chosen to limit some of the long routes generated when optimizing without energy constraints. The task locations are generated randomly and the start and terminal locations are set to the map center.

To evaluate the energy cost, we take 100 randomly distributed samples in the ground truth map, and estimate a Gaussian process map according to Sec. \ref{sec:uncertainty_model}. In practice, this sampling process should be done by scout vehicles through sensors. 
\ifremovegpr
We
\else
The parameters are (\(\sigma_f = 6, l_f = 100, \sigma_s = 1, C_\mu = 30\)), and we
\fi
scale the map value by weights to get the map for the 6 types of vehicles. With the maps, we can then generate the energy edge costs and uncertainties for the three models.

For the optimization, as time is not the primary consideration of this paper, we set the travel time between any two nodes as 1, and set the service time that each vehicle needs to spend at a task also as 1. This means, any travel or service takes 1 unit time. As there are no additional time requirements for the tasks, the schedule for a vehicle is based on energy cost minimization. The time penalty coefficient \(C_q\) is set to 1. The confidence level \(\beta_k\) in the CCP model is set to \(95\%\) for all vehicles. The recourse penalty coefficient \(C_g\) in the SPR model is set to 1.


\begin{figure}[hbt!]
	\centering
	\begin{subfigure}[b]{0.50\linewidth}
    	\includegraphics[width=1\linewidth, trim=29 195 140 205, clip]{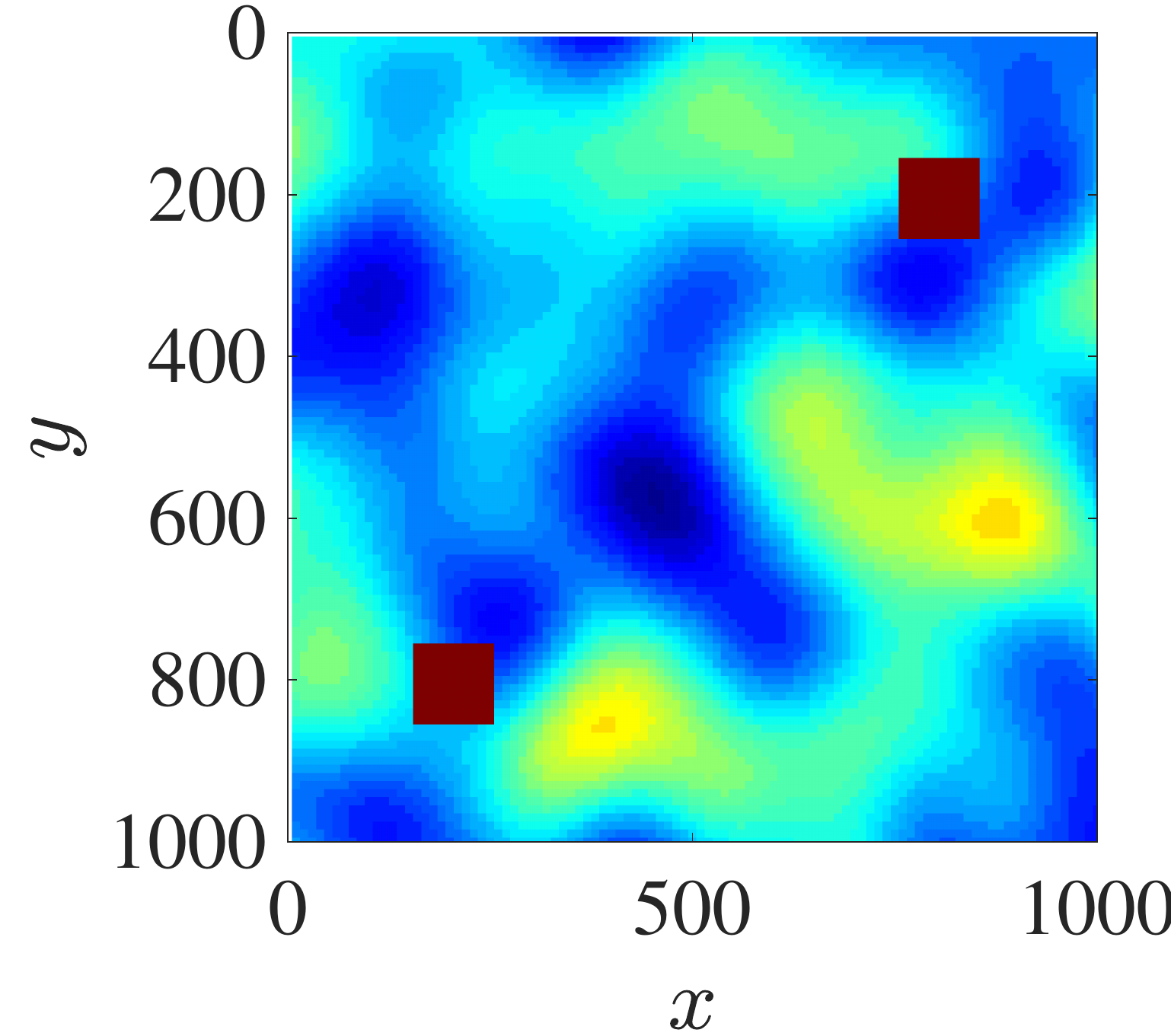}
    	\caption{Ground truth energy map.}
	\end{subfigure}
	\centering
	\begin{subfigure}[b]{0.465\linewidth}
    	\includegraphics[width=1\linewidth, trim=130 195 70 205, clip]{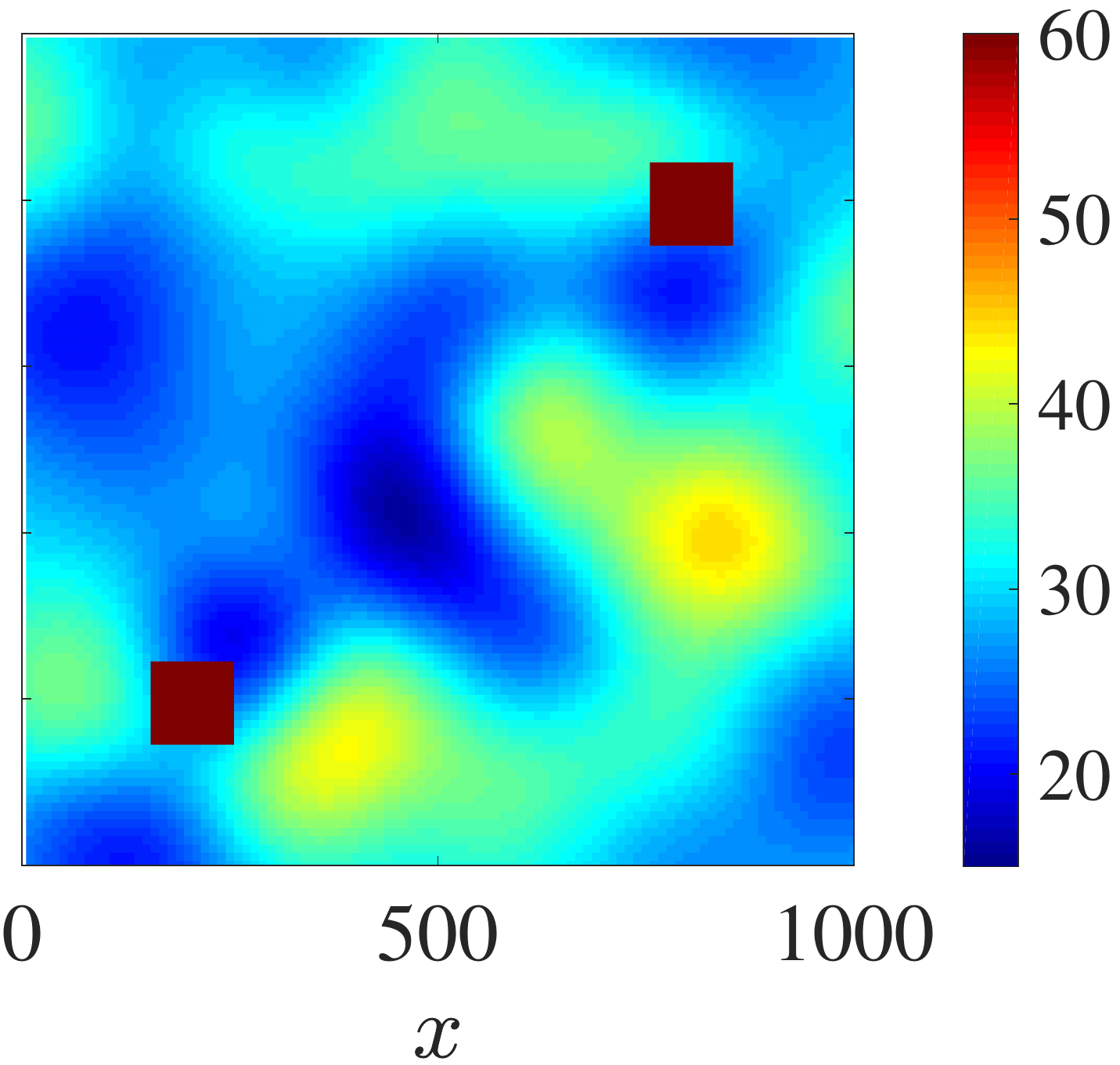}
    	\caption{Gaussian process map.}
	\end{subfigure}
	\caption{Energy maps. The red blocks are obstacles.}
	\label{fig:energy_maps}
\end{figure}

\subsubsection{Results and Discussion}

\begin{figure}[h!]
	\centering
	\begin{subfigure}[b]{0.50\linewidth}
    	\includegraphics[width=1\linewidth, trim=29 195 140 205, clip]{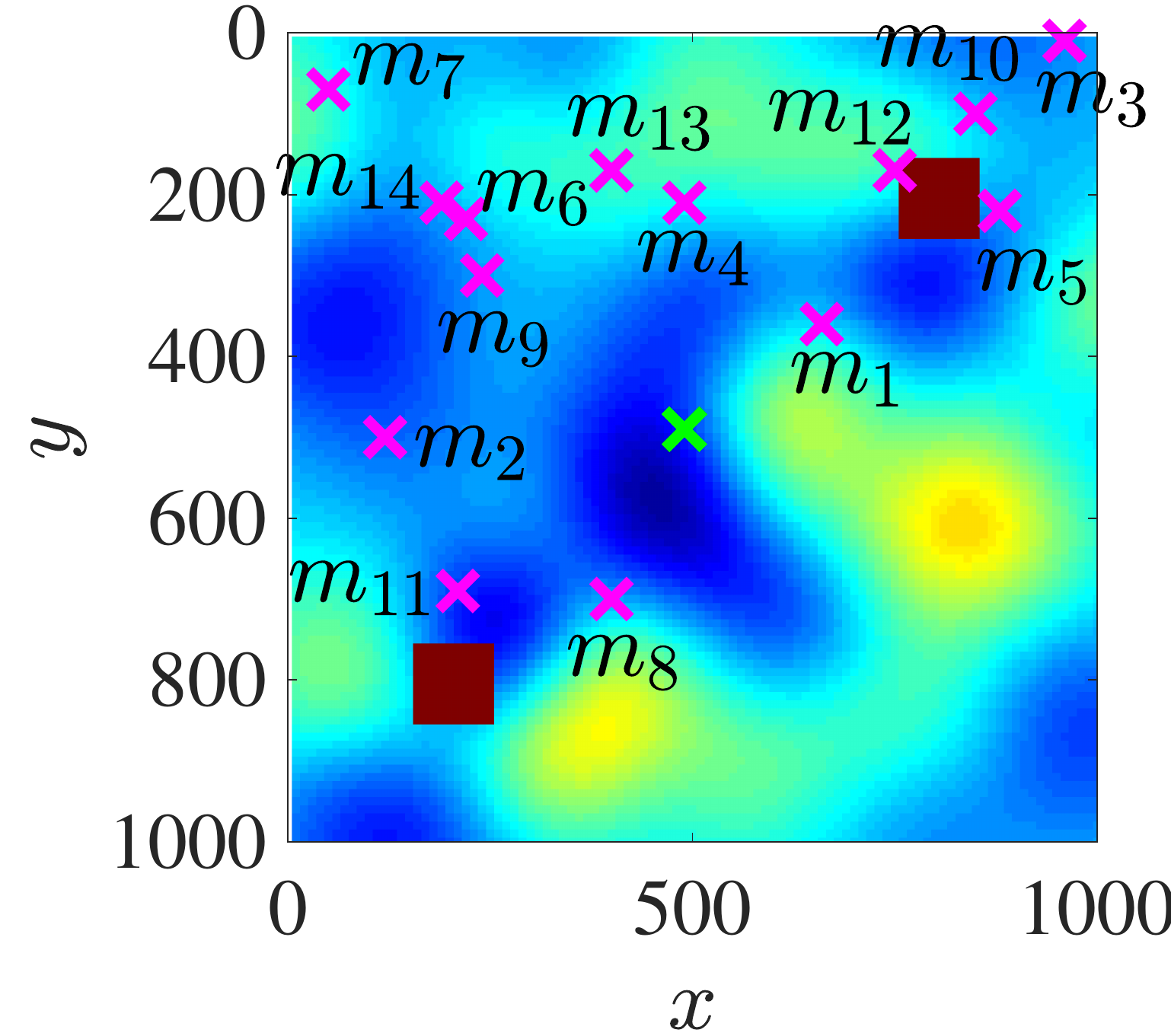}
    	\caption{Task distribution.}
	\label{fig:task_distribution}
	\end{subfigure}
	\centering
	\begin{subfigure}[b]{0.465\linewidth}
    	\includegraphics[width=1\linewidth, trim=130 195 70 205, clip]{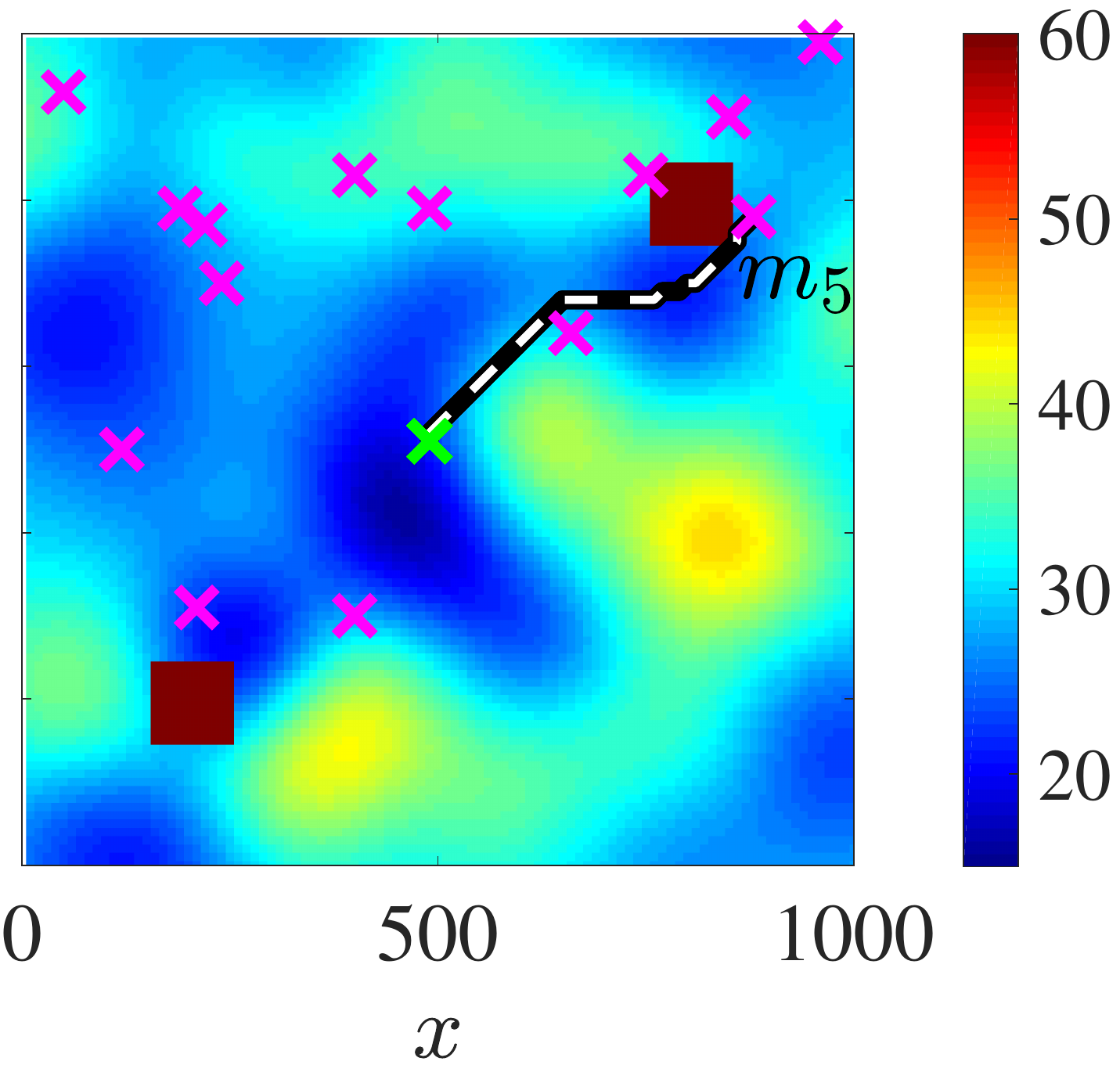}
    	\caption{CCP: route of the first \(v_5\).}
	\end{subfigure}
	\begin{subfigure}[b]{0.50\linewidth}
    	\includegraphics[width=1\linewidth, trim=29 195 140 205, clip]{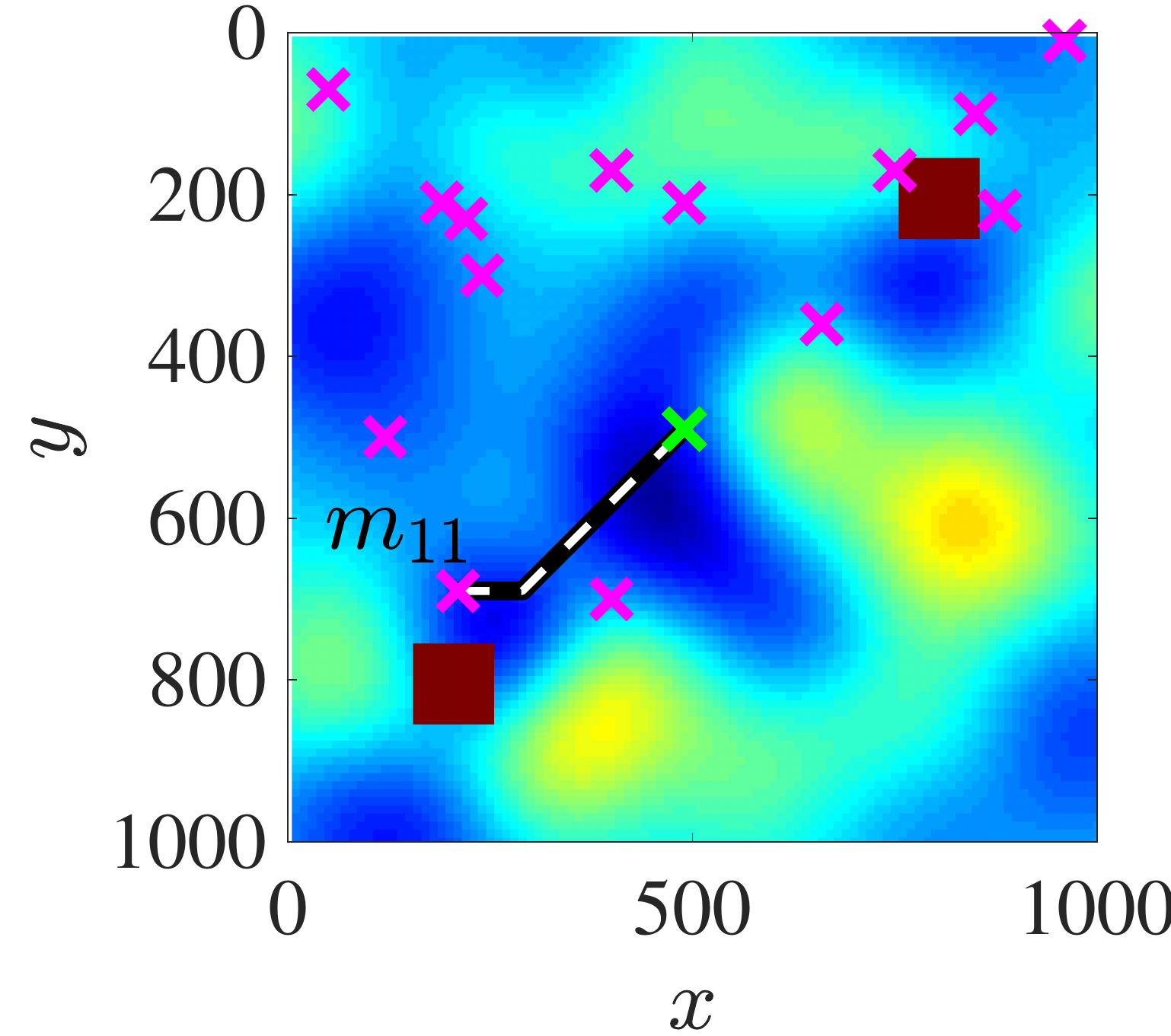}
    	\caption{CCP: route of the second \(v_5\).}
	\end{subfigure}
	\centering
	\begin{subfigure}[b]{0.465\linewidth}
    	\includegraphics[width=1\linewidth, trim=130 195 70 205, clip]{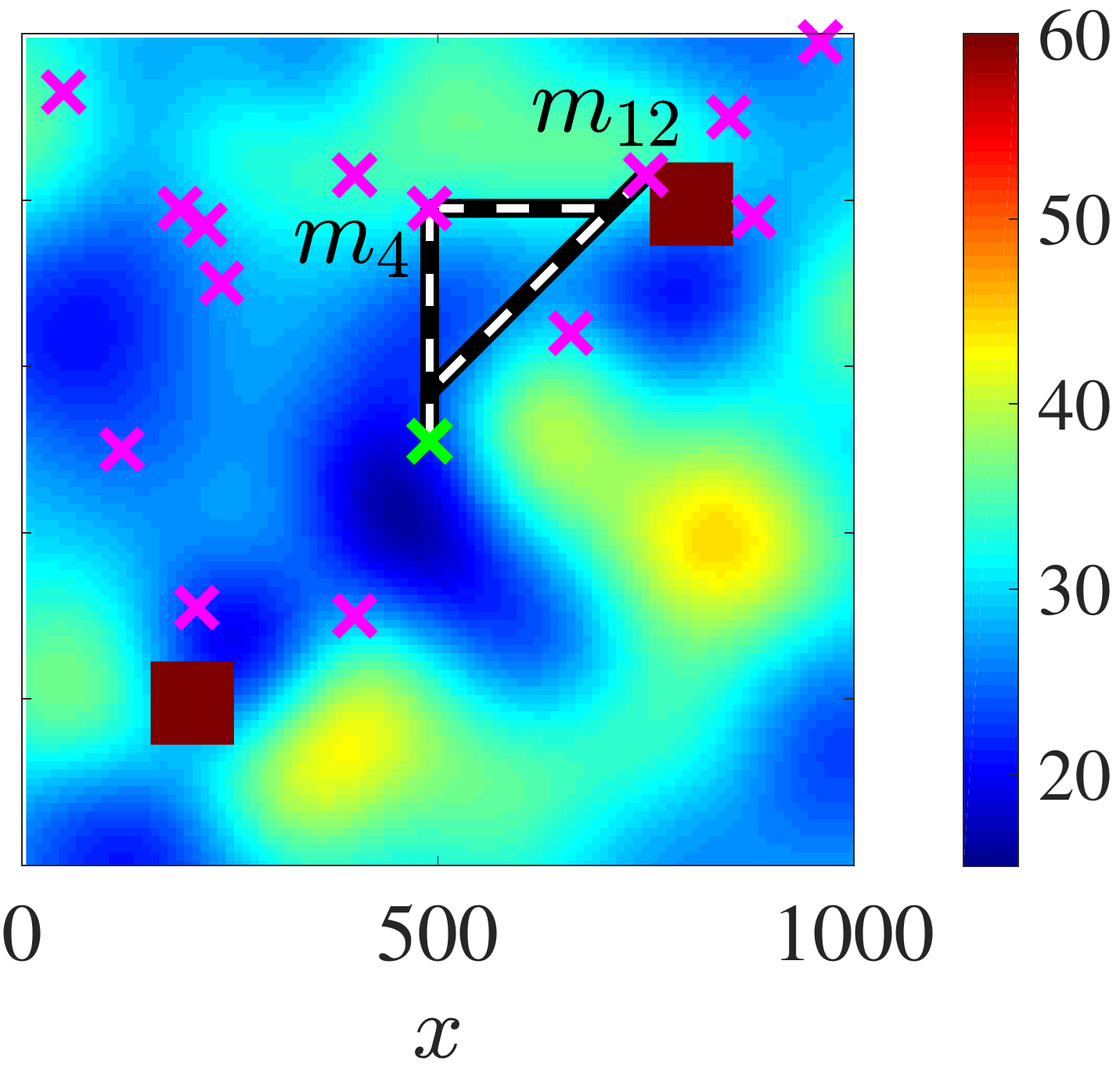}
    	\caption{CCP: route of the third \(v_5\).}
	\end{subfigure}
	\begin{subfigure}[b]{0.50\linewidth}
    	\includegraphics[width=1\linewidth, trim=29 195 140 205, clip]{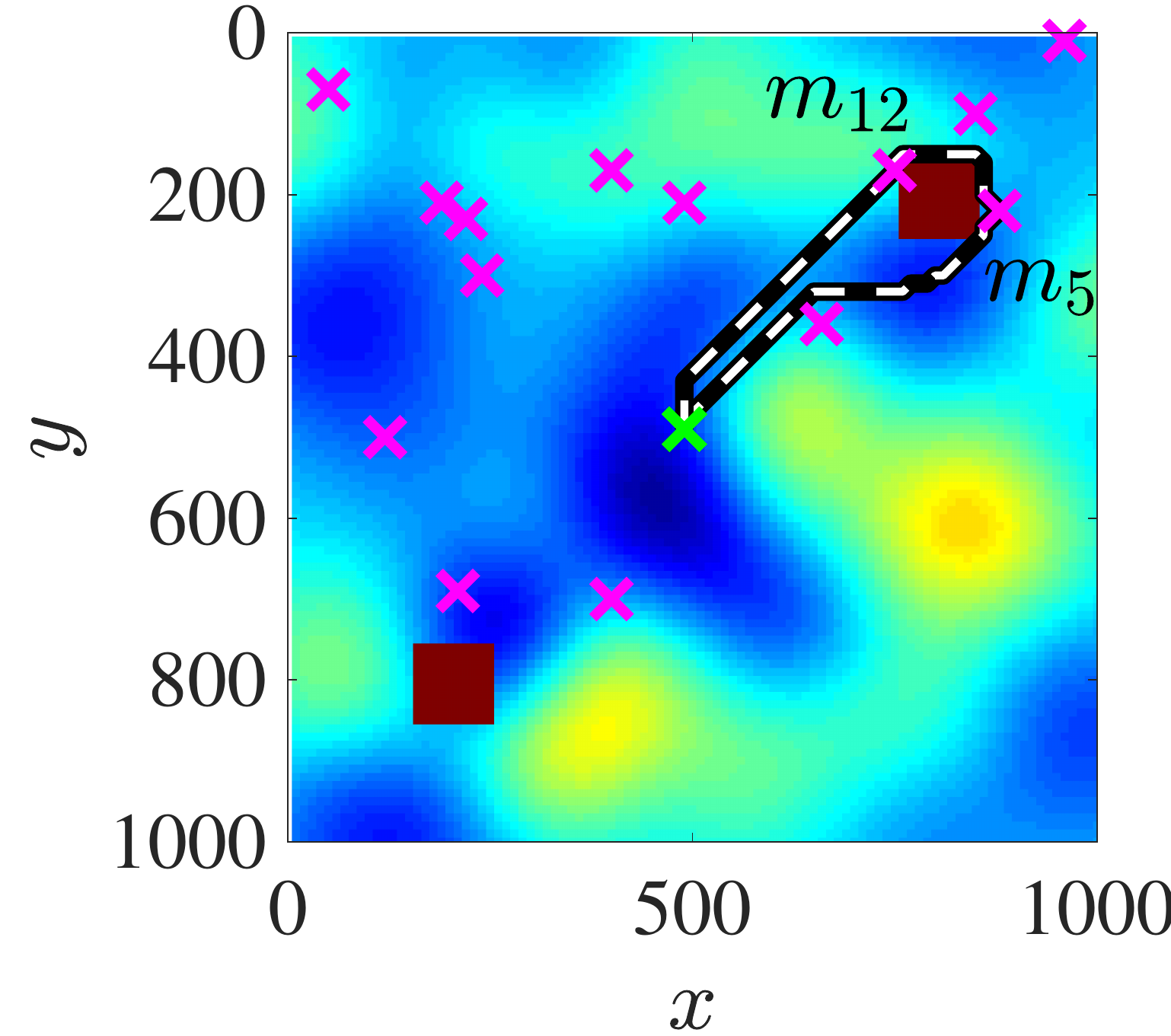}
    	\caption{SPR: route of the first \(v_5\).}
	\end{subfigure}
	\centering
	\begin{subfigure}[b]{0.465\linewidth}
    	\includegraphics[width=1\linewidth, trim=130 195 70 205, clip]{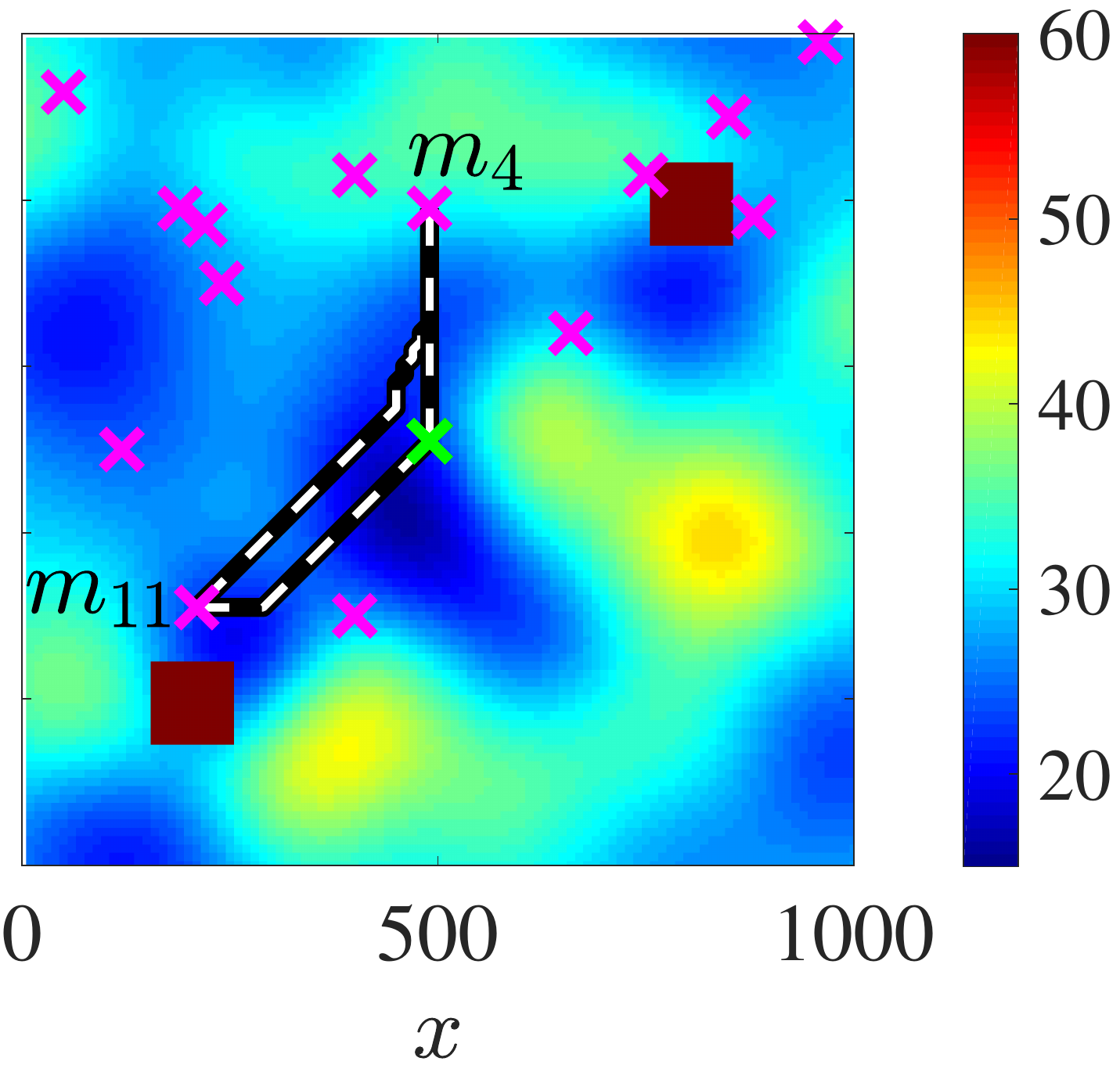}
    	\caption{SPR: route of the second \(v_5\).}
	\end{subfigure}
	\caption{Planned vehicle routes.}
	\label{fig:planning_result}
	\squeezeup\squeezeup
\end{figure}

The ground truth and estimated energy maps are shown in Fig. \ref{fig:energy_maps}. The mean and maximum errors of the estimation are 1.13 and 9.29. The randomly generated task locations are shown in Fig. \ref{fig:task_distribution}.

{\color{black} The team selection, vehicle routes, and schedules are generated. The CCP and SPR models select the same team configurations for all the tasks (TABLE \ref{tab:task_team_allocation}). Differences in the routes show that CCP generates more conservative and balanced plans than SPR.}
For instance, 
three \(v_5\) in the CCP plan are used to complete the same number of tasks as compared to two \(v_5\) in the SPR plan. The CCP routes are shorter as can be seen in Fig \ref{fig:planning_result}.

The objective values are \(f^*_\text{CCP} = 5.01\times10^6\), \(f^*_\text{SPR} = 4.78\times10^6\), and \(g^*_\text{SPR} = 5.31\times10^3\). The computation times are 28.4 and 70.6 seconds, respectively, for CCP and SPR. The small recourse cost \(g^*_\text{SPR}\) is due to the small possibility that a failure occurs. This is a sign that the CCP model might result in over conservatism.

From the team selection in TABLE \ref{tab:task_team_allocation}, we see that for all of the tasks, the requirements are fulfilled. For both models, two \(v_4\) rather than one \(v_3\) are chosen for \(m_5\) due to the lower weight (unit energy cost) of \(v_4\). However, though \(v_1\) is heavier than \(v_2\), it is used for \(m_8\) because of routing issues.

\begin{table}[h!]
	\centering
		\begin{tabular}{l|p{0.32\linewidth}|p{0.32\linewidth}}
			\toprule
			  &CCP                    & SPR \\
			\hline
			\(m_1\)&
\(v_{2}\) &
\(v_{2}\) \\ 
			\(m_2\)&
\(v_{2}\) &
\(v_{2}\) \\ 
			\(m_3\)&
\(v_{4}\) &
\(v_{4}\) \\ 
			\(m_4\)&
\(v_{1}\), \(v_{5}\), \(v_{6}\) &
\(v_{1}\), \(v_{5}\), \(v_{6}\) \\ 
			\(m_5\)&
\(v_{4} \times 2\), \(v_{5}\) &
\(v_{4} \times 2\), \(v_{5}\) \\ 
			\(m_6\)&
\(v_{3}\) &
\(v_{3}\) \\ 
			\(m_7\)&
\(v_{1}\) &
\(v_{1}\) \\ 
			\(m_8\)& 
\(v_{1}\) &
\(v_{1}\) \\ 
			\bottomrule
		\end{tabular}
	\caption{Vehicle teams for each task. The vehicle types in teams for tasks \(m_9\)-\(m_{14}\) are the same as \(m_2\)-\(m_7\).}\label{tab:task_team_allocation}
\end{table}

By comparing task locations in Fig. \ref{fig:task_distribution}, 
we see the algorithm tends to allocate the tasks that are close to each other in the map to the same set of vehicles, which is an expected behavior to reduce the total energy cost. For example, in Fig. \ref{fig:planning_result}e-f, \(m_5\) is grouped with \(m_{12}\) rather than \(m_4\) or \(m_{11}\).

As a whole, the results show that globally, the program is optimizing the routing plan considering spatial distribution, heterogeneity, and uncertainty simultaneously, while locally, the decision of one single task or vehicle might be dominated by a few factors, such as weight and task locations in previous examples.

\section{Conclusions and Future Work}\label{sec:conclusion}

We present an optimization framework that deals with the heterogeneous vehicle teaming problem with uncertain travel energy costs. The vehicle and task heterogeneity are encoded as constraints in the program. The energy uncertainty in the environment is first captured by a Gaussian process map and then turned into Gaussian distributed path costs. The uncertain path costs are either used to form probabilistic constraints in a CCP model or additional recourse costs in an SPR model. We use two branch and cut algorithms to solve the proposed stochastic programs.

Extensive computational experiments are conducted to evaluate the problem size and heterogeneity level range that the algorithm can handle, as well as the typical behaviors of the CCP and SPR models under different uncertainty levels. An explore and breach mission is shown to demonstrate one practical application of the algorithm. Results show that our algorithms can handle problems with up to 30 heterogeneous tasks and 50 heterogeneous vehicles, given the time limit is 500 seconds.

In future work, we intend to model other uncertainty categories (e.g. time uncertainty) such that the plan is robust to multiple uncertain factors in the environment. Additionally, we aim to implement and validate the proposed framework on real vehicle systems.

\section*{Acknowledgments}

DISTRIBUTION A. Approved for public release; distribution unlimited (OPSEC 4396). This research has been supported by the Automotive Research Center.




\bibliographystyle{IEEEtran}
\bibliography{ms}

\end{document}